\documentclass[12pt]{article}
\usepackage[utf8]{inputenc}
\usepackage[OT1]{fontenc}
\usepackage{charter}

\usepackage{authblk}
\usepackage{tabularx}

\usepackage[french,english]{babel}
\usepackage{hyperref}
\usepackage{longtable}

\title{Stylometry for Noisy Medieval Data:
Evaluating Paul Meyer’s Hagiographic Hypothesis}

\author[1]{Jean-Baptiste Camps}
\author[2]{Thibault Clérice}
\author[3]{Ariane Pinche}
\affil[1,2,3]{École Nationale des Chartes, PSL Research University}
\affil[2,3]{Université Lyon 3}

\date{December 2020}

\usepackage{natbib}
\usepackage{graphicx}

\begin{document}

\maketitle

\begin{abstract}
Stylometric analysis of medieval vernacular texts is still a significant
challenge: the importance of scribal variation, be it spelling or more
substantial, as well as the variants and errors introduced in the
tradition, complicate the task of the would-be stylometrist. Basing
the analysis on the study of the copy from a single hand of several
texts can partially mitigate these issues (Camps \& Cafiero, 2013), but
the limited availability of complete diplomatic transcriptions might
make this difficult. In this paper, we use a workflow combining
handwritten text recognition and stylometric analysis, applied to the
case of the hagiographic works contained in MS BnF, fr. 412. We seek to
evaluate Paul Meyer's hypothesis about the constitution of groups of
hagiographic works, as well as to examine potential authorial groupings
in a vastly anonymous corpus.
\end{abstract}

\section[Introduction]{Introduction}\label{introduction}

\subsection[Understanding the French Hagiographic Tradition]{Understanding the French Hagiographic Tradition}
\label{understanding-the-french-hagiographic-tradition}

The history of the early French saint's Lives collections in prose is
still a mystery. Indeed at the beginning of the thirteenth century,
\emph{legendiers} (i.e. manuscript containing Saint's Lives collection)
were already constituted and preliminary steps are missing. In the case
where those collections do not adopt the liturgical calendar, they often
are built around thematic series (Perrot, 1992:11-15): apostles,
martyrs, confessors, saint virgins, but the organization within those
themes isn't clear. One of the hypotheses about the composition of
hagiographic collections in Latin with similar structure is that they
are a compilation of pre-existing \emph{libelli}, independent units
about one saint or a series of saints (Philippart, 1977).

Paul Meyer's\footnote{%
Paul Meyer (1840-1917) is one of
  the most famous French romance philologists. He's the author of the
  chapter about French hagiographic Lives in the book \emph{Histoire
  littéraire de la France} (1906).%
  } work on the composition of
Old French prose legendiers (Meyer, 1906) led him to discover that some
of these came from successive compilations. Using their macrostructure,
he tried to organize French manuscripts into families. The first three
collections named A, B and C are composed by successive additions.
Thereby, collection A is a collection of saint apostles' Lives,
collection B adds a collection of saint martyrs' Lives and collection C
is the aggregation of collection A, B and 22 new texts: saint
confessors' Lives, saint virgins' Lives, one text about the antichrist
and another about the purgatory. New additions to the collection are not
united by a thematic object and seem messier. Studying those
compilations, Paul Meyer had also the intuition that the collections
possess some smaller pieces. He identified a few series using authorship
when he could (eg. \emph{Li Seint Confessor} of Wauchier de Denain in
collection C) and proposed the existence of primitive series based on
the repetitive grouping of selected lives in different manuscripts as
for instance the series: Saint Sixte, Saint Laurent and Saint Hippolyte.

\begin{figure}
    \centering
    \includegraphics[width=\textwidth]{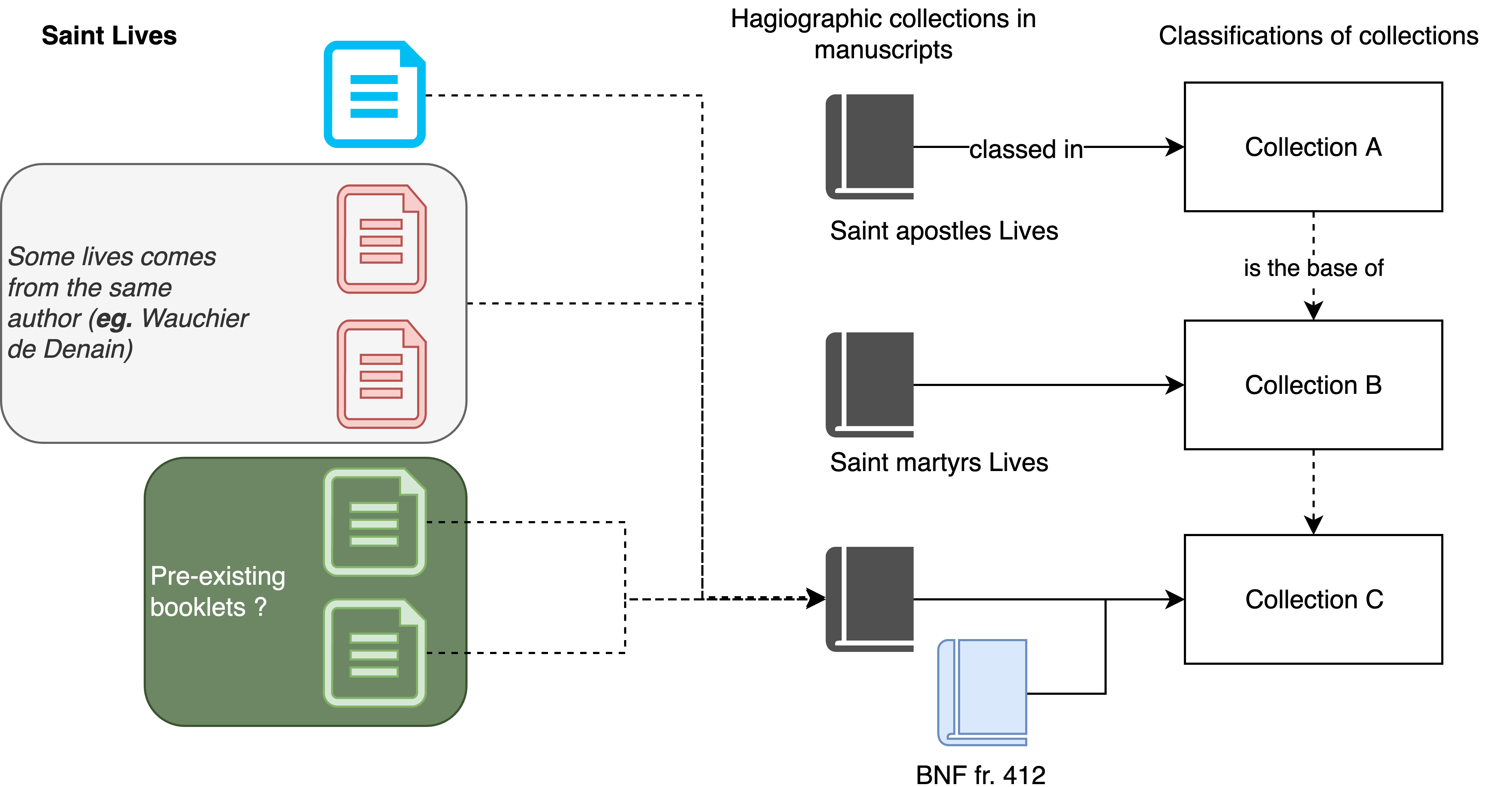}
    \caption{Representation of the possible composition of French
hagiographic collections based on Paul Meyer's hypothesis}
    \label{fig:image1}
\end{figure}

Because most of the French saint's lives are anonymous and because the
collections were rearranged by multiple editors over time, it is
extremely difficult to locate what could have been the primitive series
and Meyer couldn't go further. This serial composition of the Lives of
Saints is a datum also noted by other specialists of Latin hagiography
such as Perrot (1992) and Philippart (1977), who even points out that
these hagiographic series must be studied in their entirety in the same
way as a literary work. However, despite these academic positions, to
our knowledge, there is no complete edition of hagiographic series, most
probably because the identification of the series themselves is a matter
of debate, except in the context of a full manuscript edition.

As such, the aim here is, first, to determine if Paul Meyer's intuitions
and hypotheses can be infirmed, nuanced or completed. Second, we would
like to discover if some other links between saint's lives can reveal
series from single anonymous authors and help reconstitute some of the
hypothetical pre-existing \emph{libelli}. In order to do so, we
performed a stylometric analysis on one representative of the collection
C. To perform this computational approach, and because there is no
complete edition of any of the manuscripts holding the collection C, we
created a pipeline to acquire the text. After the presentation of the
data acquisition pipeline, we will explain how we approach the inherent
problems of both Old French and OCR variability in our stylometric
analysis. Finally, we'll propose an evaluation of the results in regard
to the traditional knowledge we have of the manuscript transmissions.

\section[Development and Evaluation of a Data
Pipeline]{Development and Evaluation of a Data Pipeline}
\label{development-and-evaluation-of-a-data-pipeline}

For this work, the BnF fr. 412 manuscript, written in a single hand
during the 13th century, seems to be a valid source for text
acquisition: it was most probably written in a short time, which does
not impact too much writing style (the writer's hand does not ``age''),
it is in pristine condition and has been digitized by the Bibliothèque
Nationale de France (BNF) and made available on
Gallica\footnote{Manuscript BnF fr. 412 on Gallica :
  \textless{}\url{https://gallica.bnf.fr/ark:/12148/btv1b84259980/}\textgreater{}.}.
To be able to analyze the text, we had to build a pipeline that would,
step by step, enrich the data with more information: from pictures to
text, from raw text to normalized version, from normalized version to
linguistically annotated data so that multiple stylometrical approaches
could be combined and evaluated.

\subsection[Line Detection]{Line Detection}
\label{line-detection}

Handwritten Text Recognition has evolved much over the course of the
past years, with easy to use tools such as Transkribus and Kraken. We
distinguish two steps of text acquisition: layout detection (and
particularly line detection) and the actual text recognition.

As Garz \emph{et al.} (2012) put it, ``segmenting page images into text
lines is a crucial pre-processing step for automated reading of
historical documents" : unlike printed books from modern editions,
parchments present various issues from ink bleed-through (the capacity
of a verso writing or picture to be seen on a recto) to inconsistent
background color. On top of these traditional issues, costly manuscripts
like the BnF fr. 412 accompany texts with illumination, including
historiated lettrines, flourished initials and marginal ornamentation,
as well as rubrics, underscoring in this way the discontinuity between
texts (see Fig. 2).

\begin{figure}
    \centering
    \includegraphics[width=\textwidth]{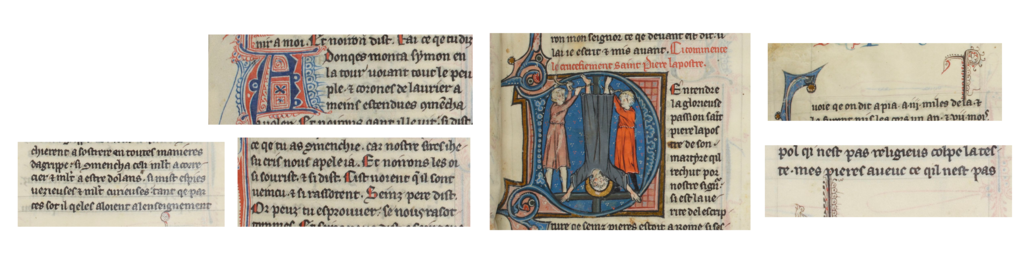}
    \caption{Examples of manuscript layout issues for line
segmentation: ink bleed-through and disappearance of text (bottom left),
flourished and highlighted initials (second column from the left, both
top and bottom), historiated initials (third column), litterae elongatae
on first and last lines (last column) from fol.9r and
fol.10r}
    \label{fig:image2}
\end{figure}

Line detection was found to perform very poorly in Kraken compared to
Transkribus, two well known and performant HTR and OCR engines. Kraken
in its 2.0.5 release contains a traditional line segmenter based on
contrast which cannot be trained on a specific layout while Transkribus
is using deep-learning models for the same
work\footnote{It should be noted that Kraken main
  developer, Benjamin Kiessling, is working on implementing learnable
  layout analysis (cf.
  \href{https://github.com/mittagessen/kraken/issues/155}{\emph{https://github.com/mittagessen/kraken/issues/155}}).%
  }. While it should be stressed that we cannot offer a methodical,
reapplicable evaluation for this performance, we can definitely say that
Kraken would often miss lines, create a lot of false positives in
ornamentation, and - not often but enough to be seen - incorrectly sort
the lines. On the other hand, Transkribus would rarely miss lines,
rarely find text in illuminations (although it could happen), but had
some time issues with last lines of columns (see Fig. 3). We believe in
subsequent results Kraken output to be much noisier than Transkribus.

\begin{figure}
    \centering
    \includegraphics[width=1.57153in,height=1.89028in]{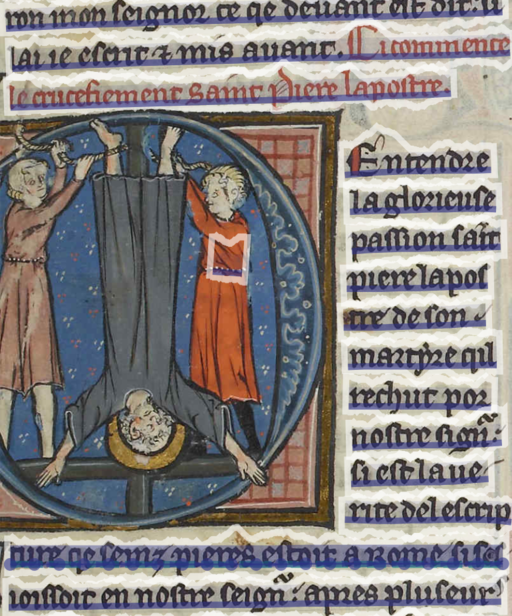}\includegraphics[width=2.18472in,height=0.54653in]{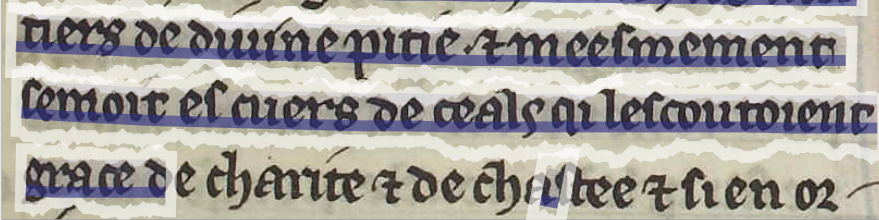}\includegraphics[width=1.64792in,height=1.91042in]{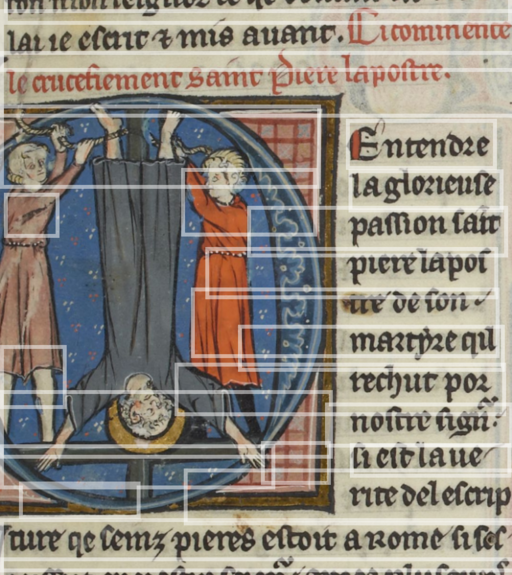}
    \caption{Line detection in both Transkribus (first and second from
the left) and Kraken. White areas are marked as lines. In these
examples, we can see the illumination incorrectly identified as text
(much more important in Kraken that in Transkribus), the false negative
on the last line in Transkribus and various issues in Kraken in
general.}
    \label{fig:image3}
\end{figure}

\subsection[HTR]{HTR}
\label{htr}

Text acquisition was evaluated using both Transkribus (HTR+) and Kraken.
Two datasets have been created for these specific reasons:

\begin{enumerate}
\item
  The main dataset, Pinche Dataset below, is the combination of 271
  columns transcribed by A. Pinche, spanning from folio 103r to folio
  170v \emph{(Pinche, }In progress\emph{)}. It has the advantage of
  having only one transcriber and has been proofread in the context of
  an ongoing PhD thesis. It contains 96 characters (single spaces
  included), of which 28 are found fewer than 10 times, and in total
  makes up to around 495,000 occurrences. However, it has the downside
  of being both consecutive and attributed to a single author (Wauchier
  de Denain).
\item
  The second dataset, below TNAH Dataset, is the combination of 43
  columns transcribed by S. Albouy, C. Andrieux, H. Dartois, M. Frey, O.
  Jacquot, M. Morillon, M.-C. Schmied and L. Vieillon, students of A.
  Pinche in the context of her TEI course (Pinche et
  al., 2019). Unlike the Pinche Dataset, it is neither consecutive nor
  attributed to the same legendaries or authors, in fact, all of them
  are anonymous. They are composed of the \emph{Vie de Saint Philippe}
  (45ra-45vb), \emph{Vie de Saint Jacques le Mineur} (45vb-46vb), part
  of \emph{Vie de Saint Longin} (51rb-52vb), \emph{Vie de Sainte Lucie}
  (71ra-72va), \emph{Vie de Saint Sixte} (87ra-88va), \emph{Vie de
  Sainte Marguerite} (213ra-214rb), \emph{Vie de Sainte Pélagie}
  (214rb-215rb), \emph{Vie de Saint Eufrasie} (224vb-225vb). The
  downside of this dataset is that it was mostly transcribed by
  non-specialists and despite several attempts to unify it still
  presents differences in how the text was transcribed. It contains 102
  different characters (single spaces included), of which 46 are found
  less than 10 times, and in total makes up to around 70,000
  occurrences.
\end{enumerate}

\begin{figure}
    \centering
    \includegraphics[width=0.4\linewidth]{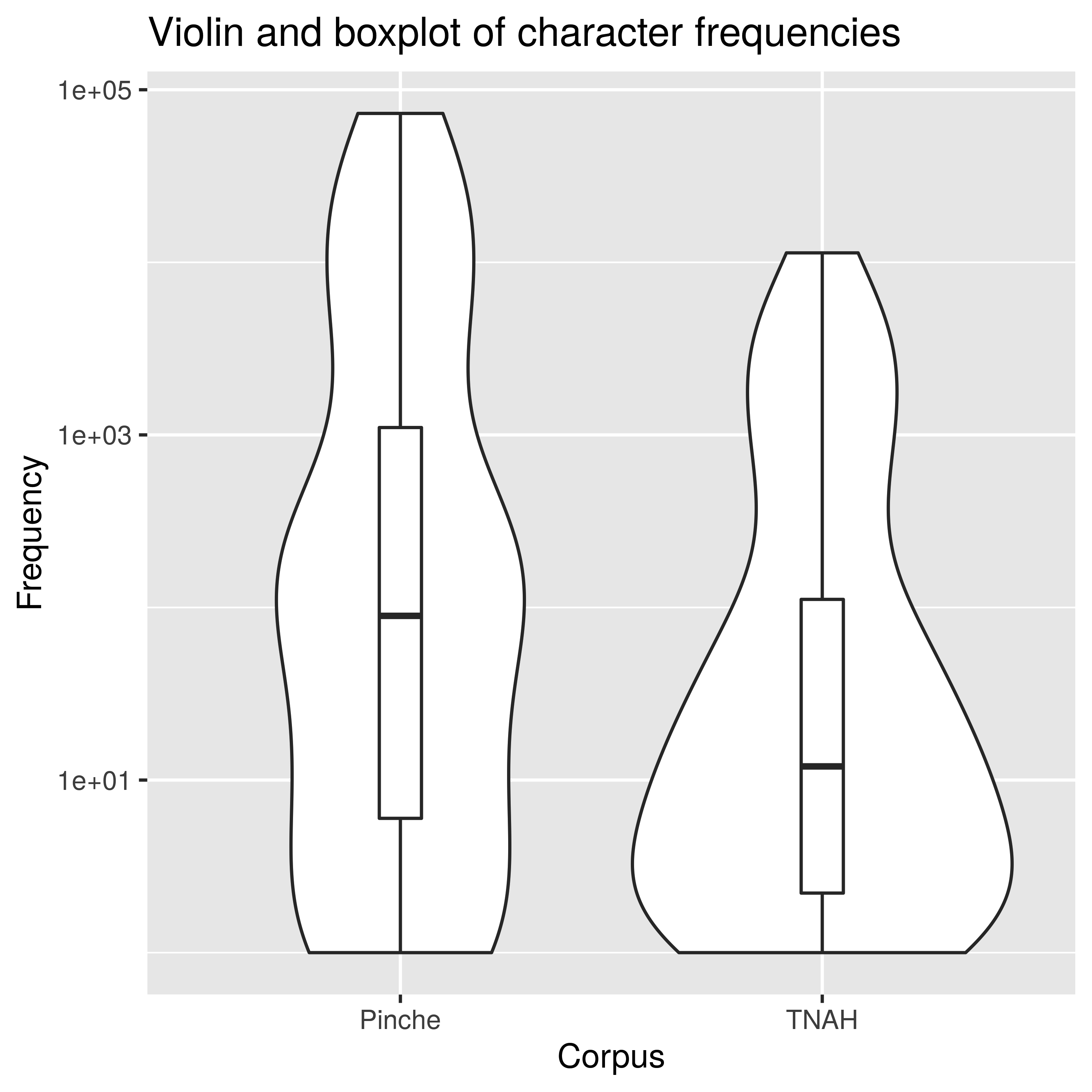}
    \caption{Distribution of character counts}
    \label{fig:image4}
\end{figure}

We trained 3 models, each of them tested on the same subset of Pinche
dataset. As expected, the training set from Pinche was more efficient
(most probably due to its single expert transcriber). However, we found
Kraken to be quite impacted by the recognition of spaces. As such, we
trained a supplementary second model that would not try to recognize
spaces. The Transkribus HTR+ model performed best on the Character Error
Rate (\emph{see Table 1}).

\begin{table} 
\small
\centering
\begin{tabular}{l|l|ccc} 
 &  & \multicolumn{3}{c}{Engine} \\ 
Test set & Training set & Kraken & Kraken (No Space) & Transkribus (HTR
+)\\
\hline \hline
PINCHE & PINCHE & 4.87 & 3.37 & 2.29\\
TNAH & PINCHE & 31.16 & 27.16 & 8.07\\
PINCHE & TNAH & N/A & N/A & 4.97
\end{tabular}
\caption{Character error rate based on software and corpora}
\label{table:1}
\end{table}

Folio 1r-3v were excluded from OCR, because they contain unrelated
resources (mostly calendars).

\subsection[Word Segmentation]{Word Segmentation}
\label{word-segmentation}

As it can be seen in the resulting text (see Fig. 5), spaces are one of
the least stable features to be correctly recognized. If spacing in
handwriting is rarely really regular, Old French manuscripts are a prime
examples of it\footnote{However, it is not a specific
  handwriting issue, but more a question of what was space and words at
  the time. On this subject, see
  Stutzmann (2019).
  Moreover, there is a strong difficulty in appreciating the difference
  between the intent of the scribe and what we perceive of it (Careri et
  al., 2001: XXXVII).}. Indeed, a quick look at one column from the
f.10r in Figure 5 shows that spaces are sometimes really small,
sometimes non-existent. Moreover, there are no marks for hyphenation in
the manuscript, which requires us to detect and concatenate some of the
tokens passing from one line to another. As an indication, the Kraken
model trained with spaces had 905 errors related to spaces from which
810 were deletions and insertions: it represents a 1.82 point drop of
performance in CER and an impressive 37.39\% of the test set errors.

\begin{figure}
    \centering
    \includegraphics[width=0.2\textwidth]{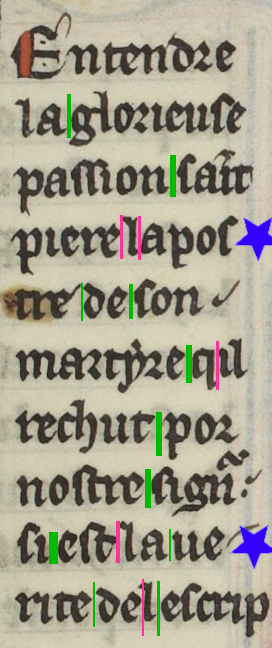}
    \caption{Part of f.10ra. Green lines are measurable space, pink are
spaces that are not written, stars represent hyphenation. Transcription
with space width in square brackets (for reference, first ``a'' width is
31px wide), ± 2px: ``Entendre la{[}4{]}glorieuse passion{[}6{]}saint
pierre{[}0{]}l{[}0{]}apos tre{[}2{]}de{[}4{]}son
martyre{[}6{]}q{[}0{]}il rechut{[}6{]}por nostre{[}6{]}sigñ
si{[}9{]}est{[}0{]}la{[}2{]}ve rite{[}2{]}de{[}0{]}l{[}3{]}escrip''.
Largest space in this area represents 35\% of the reference a, while
smallest measures 6.45\% of it. A space is the area available between
the furthest right pixel of the left letter and the leftmost one of the
next.}
    \label{fig:image7}
\end{figure}

An option is to treat the notion of space as a natural language
processing (NLP) task, where the image is not taken into account. Of
course, the notion of words and grammar has evolved and what most of the
other tools of the pipeline expect are words perceived as such by modern
and contemporaneous medievalists. Unfortunately, due to the extreme
variation in spelling of Old French, dictionary approaches do not
perform well. In a previous study
(Clérice, 2019), we
have shown that they do not extend to new unknown domains as much as
deep learning models. In this context, we used Boudams, the tool
developed for the aforementioned paper. It currently removes all spaces
before reinserting new ones. We used the Old French model built for that
study, which had a 0.99 FScore on the in-domain test set (which
contained resources from the Pinche Dataset and TNAH Dataset) while
having a 0.945 FScore on an out of domain dataset.

\begin{table} 
\small
\centering
\begin{tabular}{p{0.3\textwidth}|p{0.6\textwidth}} \small
\textit{Transkribus} & Entendre la glorieuse passion saĩt aun piere lapos stre de
son. " mart\.{y}re qil rechut por nostre sign: si est la ue rite del escrip
ture\tabularnewline \hline
\textit{Transkribus}\\ + \textit{Boudams} & Entendre la glorieuse passions aĩt a un piere la
posstre de son ." mart\.{y}re qil rechut por nostre sign : si est la uerite
del escripture\tabularnewline \hline
\textit{Kraken (No Space)}\\
+ \textit{Boudams} & Entendre la glorieuse passions a\textsuperscript{a}t piere la
positre de son . mant\.{y}re qil rechut por nostre sign : si est la uerite
del escriptire\tabularnewline \hline
\textit{Correct} & Entendre la glorieuse passion saït piere l apostre de son
mart\.{y}re q il rechut por nostre sign\textasciitilde{} si est la uerite de
l escripture\tabularnewline
\end{tabular}
\caption{Example of results and ground truth after and before word
segmentation}
\label{table:2}
\end{table}

Of course, the resulting output is not expected to be perfect (see Table
2), and in fact, each step we pursue might introduce new errors as they
were not manually transcribed or corrected. However, we did keep the
output of each step for later stylometric analyses.

\subsection[Abbreviation resolution, normalization and
lemmatization]{Abbreviation resolution, normalization and lemmatization}\label{abbreviation-resolution-normalization-and-lemmatization}

With word segmentation available, there were two others forms of the
dataset that were needed: one where each word would be normalized and
have its abbreviations resolved, a second where each word would be
tagged with both its Part-Of-Speech and its lemma. We actually treated
normalization and abbreviation resolution as a lemmatization task, as
they both require understanding of phenomenon such as prefix and suffix
and replace them with a neutral value.

As such, we trained Pie
(Manjavacas et al.,
2019) on a corpus of Old French transcriptions available in
TEI\footnote{We used a model on version 0.2.3 (
  \href{https://github.com/emanjavacas/pie/releases/tag/v0.2.3}{\emph{https://github.com/emanjavacas/pie/releases/tag/v0.2.3}}
  ).}. The training set was composed of around 125,000 tokens
(including punctuation), the evaluation set 16,000 and the test set
15,000 taken from both Pinche and Oriflamms project (Stutzmann et al.,
2013). They contained abbreviation resolution, accentuation and
punctuation introduction (sen -\textgreater{} s'en). The results were
promising with 96.86\% accuracy, with 96.96\% on ambiguous tokens (whose
input can be normalized in different fashions), 91.42 \% on unknown
output form, and finally 90.72 \% on unknown origin form.

\begin{table}
\small
\centering
\begin{tabular}{p{0.3\textwidth}|p{0.6\textwidth}} \small
\textit{Transkribus Raw} & entendre la glorieuse passion \textbf{saint} aun piere
lapos stre de son . martyre q'il rechut por nostre sign\textbf{or} : si
est la ve rite del escrip ture\tabularnewline \hline
\textit{Transkribus} + \textit{Boudams} & entendre la glorieuse passions \textbf{aint} a
un piere la posstre de son . martyre q'il rechut por nostre
sign\textbf{or} : si est la verité del escripture\tabularnewline \hline
\textit{Kraken (No Space)}\\
+ \textit{Boudams} & entendre la glorieuse passions \textbf{art} piere la positre de son .
mantyere q'il rechut por nostre sign\textbf{or} : si est la verité del
escriptire \tabularnewline \hline
\textit{Correct} & Entendre la glorieuse passion \textbf{sain}t piere l apostre
de son mart\.{y}re q il rechut por nostre sign\textbf{eur} si est la uerite
de l escripture
\end{tabular}
\caption{Example of results and ground truth after and before abbrevations resolutions resolution}
\label{table:3}
\end{table}

To improve statistical calculations based on occurrence counts, we
applied lemmatization. Unlike modern English, Old French is both defined
by its spelling variation (not only between regional \emph{scriptae} but
also inside them), and its rich morphology. As such, the same word with
different flexions can be written in different fashions. In the Pinche
Dataset, which represents 27.34\% of the whole corpus to be lemmatized
in Transkribus\footnote{In the transkribus dataset,
  Wauchier counts for 122,000 tokens over 446,900 for the whole treated
  part of the manuscript.}, the verb \emph{avoir} (to have) has 57
different spellings, the pronoun \emph{il} 17, the nouns \emph{emperëor
}8, the adverb \emph{tout} (all) 14, the adjective \emph{saint} 11: eg.
``\emph{compagnie}'' can be found written as \emph{compagnie,
compaignie, compaignies, compaigniez, conpagnie, conpaignie,
conpaignies.}

Pie is a lemmatizer specifically designed to deal with historical
languages with such traits as those found in Old French. We trained a
lemmatizer on a dataset of approximately 500,000 lemmatized tokens which
were taken from the Chrestien corpus
(Kunstmann, 2009),
the Geste corpus
(Camps, 2019), the
Institutes (Olivier-Martin et al., 2018), the Lancelot (Ing, in progress)
and the Wauchier (Pinche, In progress)
datasets\footnote{The lemmata are taken from Tobler, A.,
  and Lommatzsch, E. (1952). \emph{Altfranzösisches Wörterbuch}. E.
  Steiner.}. The overall model had 96.38 \% accuracy on the test corpus
comprised of 48,317 tokens, punctuation included
(Clérice et al.,
2019).

\begin{table}
\small
\centering
\begin{tabular}{ll|llll} \small
& & Accuracy & Precision & Recall & Support\tabularnewline \hline \hline
\textbf{Lemma} & All & 96.38 & 71.23 & 70.89 & 48,317\tabularnewline
& Ambiguous Tokens & 96.65 & 75.85 & 76.43 & 27,844 \tabularnewline
& Unknown Lemma & 72.9 & 26.85 & 26.03 & 1,236 \tabularnewline
& Unknown Form & 64.29 & 42.9 & 42.49 & 1,792 \tabularnewline
\textbf{POS} & All & 96.13 & 78.28 & 75.72 & 48,317\tabularnewline
& Ambiguous Tokens & 95.49 & 78.88 & 75.3 & 32,232 \tabularnewline
& Unknown Form & 86.77 & 59.59 & 59.18 & 1,792 \\
\end{tabular}
\caption{Test scores of Pie over Old French}
\label{table:4}
\end{table}

\begin{table}
\small
\centering
\begin{tabular}{l|p{0.7\textwidth}} \small
Transkribus Raw & entendre\textless{}VERinf\textgreater{}
le\textless{}DETdef\textgreater{} glorïos\textless{}ADJqua\textgreater{}
passïon\textless{}NOMcom\textgreater{}
amer1\textless{}VERcjg\textgreater{} a3\textless{}PRE\textgreater{}
un\textless{}DETndf\textgreater{} piere\textless{}NOMcom\textgreater{}
le\textless{}DETdef\textgreater{} postre\textless{}NOMcom\textgreater{}
de\textless{}PRE\textgreater{} son4\textless{}DETpos\textgreater{}
.\textless{}PONfrt\textgreater{} martire2\textless{}NOMcom\textgreater{}
que2\textless{}CONsub\textgreater{} '\textless{}PONfbl\textgreater{}
il\textless{}PROper\textgreater{}
recevoir\textless{}VERcjg\textgreater{} por2\textless{}PRE\textgreater{}
nostre\textless{}DETpos\textgreater{}
seignor\textless{}NOMcom\textgreater{} :\textless{}PONfbl\textgreater{}
si\textless{}ADVgen\textgreater{} estre1\textless{}VERcjg\textgreater{}
le\textless{}DETdef\textgreater{} verité\textless{}NOMcom\textgreater{}
de+le\textless{}PRE.DETdef\textgreater{}
escriture\textless{}NOMcom\textgreater{}\tabularnewline \hline
Correct & entendre\textless{}VERinf\textgreater{}
le\textless{}DETdef\textgreater{} glorïos\textless{}ADJqua\textgreater{}
passïon\textless{}NOMcom\textgreater{}
saint\textless{}ADJqua\textgreater{}
Pierre\textless{}NOMpro\textgreater{} le\textless{}DETdef\textgreater{}
apostle\textless{}NOMcom\textgreater{} de\textless{}PRE\textgreater{}
son4\textless{}DETpos\textgreater{}
martire2\textless{}NOMcom\textgreater{}
que2\textless{}PROrel\textgreater{} il\textless{}PROper\textgreater{}
recevoir\textless{}VERcjg\textgreater{} por2\textless{}PRE\textgreater{}
nostre\textless{}DETpos\textgreater{}
seignor\textless{}NOMcom\textgreater{} si\textless{}ADVgen\textgreater{}
estre1\textless{}VERcjg\textgreater{} le\textless{}DETdef\textgreater{}
verité\textless{}NOMcom\textgreater{} de\textless{}PRE\textgreater{}
le\textless{}DETdef\textgreater{}
escriture\textless{}NOMcom\textgreater{}\tabularnewline
\end{tabular}
\caption{Automatic tagging of the OCR text by Pie and ground
truth}
\label{table:5}
\end{table}

The final result is a lemmatization and pos-tagging of each document.
Error accumulation through successive post-processing steps, and noise
in the source HTR dataset leads to a dataset with varying quality,
although some parts of the document, if not most, are treated with
satisfying results.

To evaluate the impact of all pipeline steps on lemmas and POS 3-grams
frequencies, in a case where the total number of words can differ, we
evaluate the differences with the ground truth followingly,

\[
\Delta_{A,B} = \frac{%
\sum_{i=1}^n \left|tf(A_i) - tf (B_i)\right| %
}{\sum_{i=1}^n tf(A_i)}
\]

Where $tf(A_i)$ is the absolute term frequency of feature $i$ in document $A$
to be evaluated, and $tf(B_i)$ it's frequency in document $B$, the ground
truth.

We also provide the ratio of hapaxes in A compared to B and vice versa.

\begin{table} 
\small
\centering
\begin{tabular}{l|ccccc|cccc}
& \multicolumn{5}{c}{Lemmas} & \multicolumn{4}{c}{POS 3-gram} \\
& \multicolumn{3}{c}{ Delta} & \multicolumn{2}{c}{Specific entries} & \multicolumn{2}{c}{Delta} & \multicolumn{2}{c}{Specific entries} \\
Corpus & All & Function & Moisl & OCR & Gold & All & Moisl & OCR &
Gold \\ \hline \hline
Martin (29, 30) & 33.35 & 10.69 & 11.43 & 34.01 & 16.92 & 44.8 & 32.28 &
29.59 & 24.23\tabularnewline
Dialogues (31) & 29.38 & 9.77 & 9.99 & 33.01 & 19.69 & 48.38 & 35.85 &
29.84 & 23.49\tabularnewline
Brice (32) & 39.49 & 12.14 & 16.56 & 29.53 & 21.39 & 66.09 & 47.51 &
30.95 & 30.86\tabularnewline
Gilles (33) & 32.24 & 9.83 & 11.07 & 25.42 & 16.91 & 46.38 & 34.03 &
25.56 & 23.64\tabularnewline
Martial (34) & 28.26 & 7.92 & 9.68 & 37.36 & 21.63 & 50.29 & 39.09 &
30.09 & 24.83\tabularnewline
Nicolas (35, 36, 37) & 29.44 & 9.33 & 10.02 & 38.76 & 21.8 & 47.19 &
35.42 & 31.42 & 24.27\tabularnewline
Jerome (38) & 34.13 & 12.59 & 14.38 & 19.53 & 15.12 & 61.92 & 52.07 &
28.66 & 28.06\tabularnewline
Benoit (39) & 27.97 & 9.64 & 11.93 & 30.98 & 17.74 & 52.88 & 44.09 &
30.88 & 24.22\tabularnewline
Alexis (40) & 30.19 & 10.65 & 11.58 & 21.83 & 13.17 & 57.71 & 47.77 &
30.16 & 26.73\tabularnewline
Total & 27.76 & 9.02 & 9.76 & 51.57 & 29.18 & 43.46 & 34.90 & 32.65 &
21.81\tabularnewline
\end{tabular}
\caption{Difference between the Gold corpus from Pinche dataset and the
HTR results at the end of the pipeline (HTR/Boudams/Pie/Pie). We provide
deltas for all values, function lemmas, and Moisl's (cf. below). Specific entries are the
relative accumulation of frequencies of lemmas that are
found in one version of the corpus but no the other.}
\label{table:6}
\end{table}

\section[Stylometric analysis]{Stylometric analysis}
\label{stylometric-analysis}

The stylometric analysis has to address several challenges, resulting
both from the nature of the texts and from the data acquisition
pipeline: the short length and anonymity of most texts; the noise in the
authorial signal caused by successive errors or innovations in the
tradition of the texts (variants) as well as the amount of spelling
variation; the noise (and potential biases) resulting from the data
acquisition pipeline. Even though stylometric methods have shown to be
relatively resilient to a - simulated or observed - moderate amount of
noise (Eder, 2013;
Franzini et al., 2018), devising a stylometric setup to partially
eliminate or circumvent it is still likely to lead to more reliable
results.

\subsection[Unsupervised analysis of short anonymous
texts]{Unsupervised analysis of short anonymous texts}
\label{unsupervised-analysis-of-short-anonymous-texts}

The texts from the manuscript are, on average, quite short, with a
median value of 3,539 words, and extreme values of 298 and 18, 971 (see
Fig. 6). Texts that are too short create a problem of reliability, as
the observed frequencies may not accurately represent the actual
probability of a given variable's appearance
(Moisl, 2011). To
limit this issue, we removed texts below 1,000 words, a relatively low
limit when compared to existing benchmarks
(Eder, 2015, 2017),
but motivated by the necessity to not exclude too many texts.

\begin{figure}
    \centering
    \includegraphics[width=0.4\textwidth]{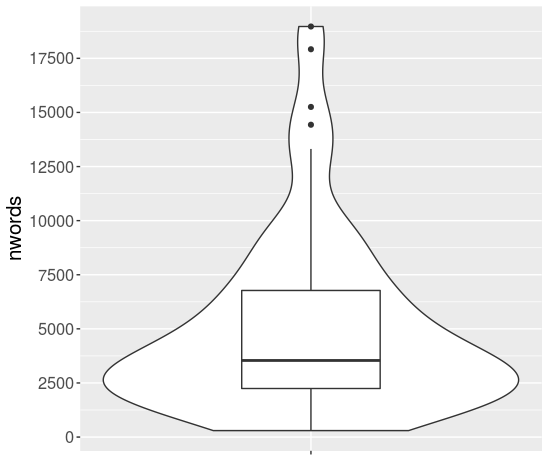}
    \caption{Boxplot of text lengths in words, based on the Transkribus
+ Boudams data}
    \label{fig:image9}
\end{figure}

Given the short length of the texts and the sparsity caused by noise, we
implement a procedure to select for analysis only those features that
satisfy a criterion of statistical reliability. In this, we follow the
procedure suggested by Moisl (2011), in the implementation already used
by Cafiero and Camps (2019). To summarize it, features are only retained
if they match the desired confidence level and margin of error even for
the smallest text in the corpus. For each feature (e.g., the function
word `et'), the minimum text size \emph{n} is calculated with

\[
n = \bar{p} \left(1 - \bar{p}\right) \left(\frac{z}{e}\right)^2
\]

where $\bar{p}$  is the mean probability of the feature in our corpus; $z$, the
confidence level, and $e$, the margin of error. We take $z = 
1.645$ to obtain a confidence margin of 90\%, and $e = 2 \sigma$,
where $\sigma$ is the feature's standard deviation. Beforehand, to
correct for normality, we generate a mirror-variable
{\emph{(Moisl, 2011)}}:

\[
vmirror_{ji} = (\max_v + \min_v) - v_{ji}
\]

where $v_j$
 is the vector of the feature $j$,
$\max_v$ and $\min_v$ are the maximum and
minimum values in $v_j$ , and $v_ji$ is the
relative frequency of $j$ in a sample $i$. This
mirror-variable is concatenated with the original variable in order to
compute $n$. If $n$ is superior to the length of the smallest
text in our corpus, we then exclude the feature from further analysis.

Because most of the texts of the manuscript are anonymous, we follow an
unsupervised approach to their analysis
(Camps and Cafiero,
2013; Cafiero and Camps,
 2019),
using agglomerative hierarchical clustering with Ward's criterion
(Ward Jr, 1963),
guided by its ability to form coherent clusters.

The metric and choices of normalization are also an important parameter,
one to which much attention has been devoted
(Evert et al., 2017;
Jannidis et al., 2015).

Following the benchmark by Evert et
al. (2017), we chose
to use Manhattan distance with z-transformation (Burrows' Delta) and
vector-length Euclidean normalization.

\subsection{Noise Reduction and Choice of
Features}\label{noise-reduction-and-choice-of-features}

In the form in which they have reached us, medieval texts are noisy by
nature, because they contain the successive errors and modifications
made by generations of scribes in the successive copies of the works.
Moreover, their language is stratified and can contain spellings and
other linguistic features originating from the dialect and regional
\emph{scripta} of the successive scribes, creating a very important and
heterogeneous spelling variation. The choice of working with the texts
of a single manuscript was already guided by the aim of limiting this
kind of noise, but is not, in itself sufficient. For this reason,
further normalisations, such as abbreviation expansion and
lemmatisation, were included in the data acquisition pipeline. Yet, even
though it achieves satisfying accuracy at each step, the pipeline
itself, through the residual presence of errors, introduces noise as
well. Moreover, since the training corpora for each algorithm were not
selected by a perfectly random process, they introduce the risk of
potential biases.

To handle these risks, we chose to retain raw as well as normalised data
for the analyses, using three feature sets:

\begin{enumerate}
\def\labelenumi{\arabic{enumi})}
\item
  \textbf{Character n-grams} from raw HTR data (baseline);
\item
  \textbf{Functors:} pseudo-affixes from expanded data, function words
  and POS n-grams;
\item
  \textbf{Words}: word forms from expanded data and lemmas.
\end{enumerate}

The aim of feature set 1 is to avoid biases resulting from the pipeline,
and for this reason to use the initial raw output of the Transkribus HTR
model, excluding all further normalisation steps. Previous research has
shown that character n-grams could be a way to circumvent issues due to
noisy OCR output, especially when compared with most frequent words
(Eder, 2013).
Following existing benchmarks
(Stamatatos, 2013),
we choose \emph{n }= 3 for our character n-grams. Because it fits our
case closely, we consider this feature set to be our baseline, and
complement it with two others.

Feature set 2 is built to capture \emph{functors}, i.e. grammatical
morphemes (Kestemont,
2014), while circumventing the noise due to scribal variation of
palaeographic and graphematic nature. Functors have long been - and
often still are - considered the most effective feature for authorship
attribution, because they capture unconscious individual variation,
while being less dependant on generic or thematic context. In this
feature set, we used expanded data to extract pseudo-affixes, i.e. a
specific kind of n-gram that has been shown, along with punctuation
n-grams, to outperform others
(Sapkota et al.,
2015), perhaps because of its ability to capture grammatical morphemes.
Since there is no authorial punctuation in our case, we extracted four
kinds of pseudo-affixes n-grams: `prefix' and `suffix' (the \emph{n}
first or last characters of words of at least n+1 characters), as well
as `space-prefix' and `space-suffix' (the interword space with the n-1
characters preceding or following it), with \emph{n} = 3. For instance,
for `\emph{annoncier}', we extracted `\^{}ann', `ier\$', `\_an' and
`er\_'. We also included function-words. Function words are commonly
recognised as one of the most effective features (if not the most) for
authorship attribution
(Argamon and Levitan,
2005; Koppel et al., 2009; Kestemont, 2014). Finally, we added
information on the morpho-syntax of the texts, by extracting
Part-of-Speech 3-grams such as `PRE DETdef NOMcom' (preposition,
definite article and noun, e.g. `\emph{a la corone}'). POS 3-grams have
sometimes shown to be a quite effective feature for cross-topic
authorship attribution
 (Gómez-Adorno et al.,
2018). In this case, multiplying the measurements by concatenating
three types of features in this set is done to help deal with short
noisy texts and improve reliability.

Feature set 3 is constituted because - despite the broad consensus on
the use of functors - some recent studies seem to advocate the use of
longer word lists as a feature for authorship attribution (Evert et al., 2017).
Using words forms is, in our case, both interesting, because it allows
us to retain morphological information, and risky, due to the extent of
spelling variation, attributable to the scribes. To account for that, we
also include lemmatized words, which, in turn, are dependent upon the
accuracy of the lemmatizer.

\subsection{Results and
Cross-Validation}\label{results-and-cross-validation}

The results on the three feature sets are included in fig.~7 HC1. Our
baseline result (see Fig.~7 HC1, top) is also the one closest to Meyer's
classification, often up to the ordering of the texts, though displaying
a few differences (6 out of 59 texts, concerning mostly texts of B
included with C). The results on feature sets 2 and 3, though keeping
the same macrostructure, display some interesting variations with the
inclusion of a mixed B/C subgroup within Meyer's A.

In order to get more insight into feature sets 2 and 3, we also give
supplementary results on their components (see Fig. 8 HC2). This can be
useful since differences on clusterings based on separate aspects (e.g.
morpho-syntactic sequences vs. function words or affixes) could reflect
differences in groupings when alternative perspectives are taken on the
language; or punctually yield useful information on some texts, as we
vary the lens with which we observe it.

\begin{figure}
    \centering
    \includegraphics[width=0.8\textwidth]{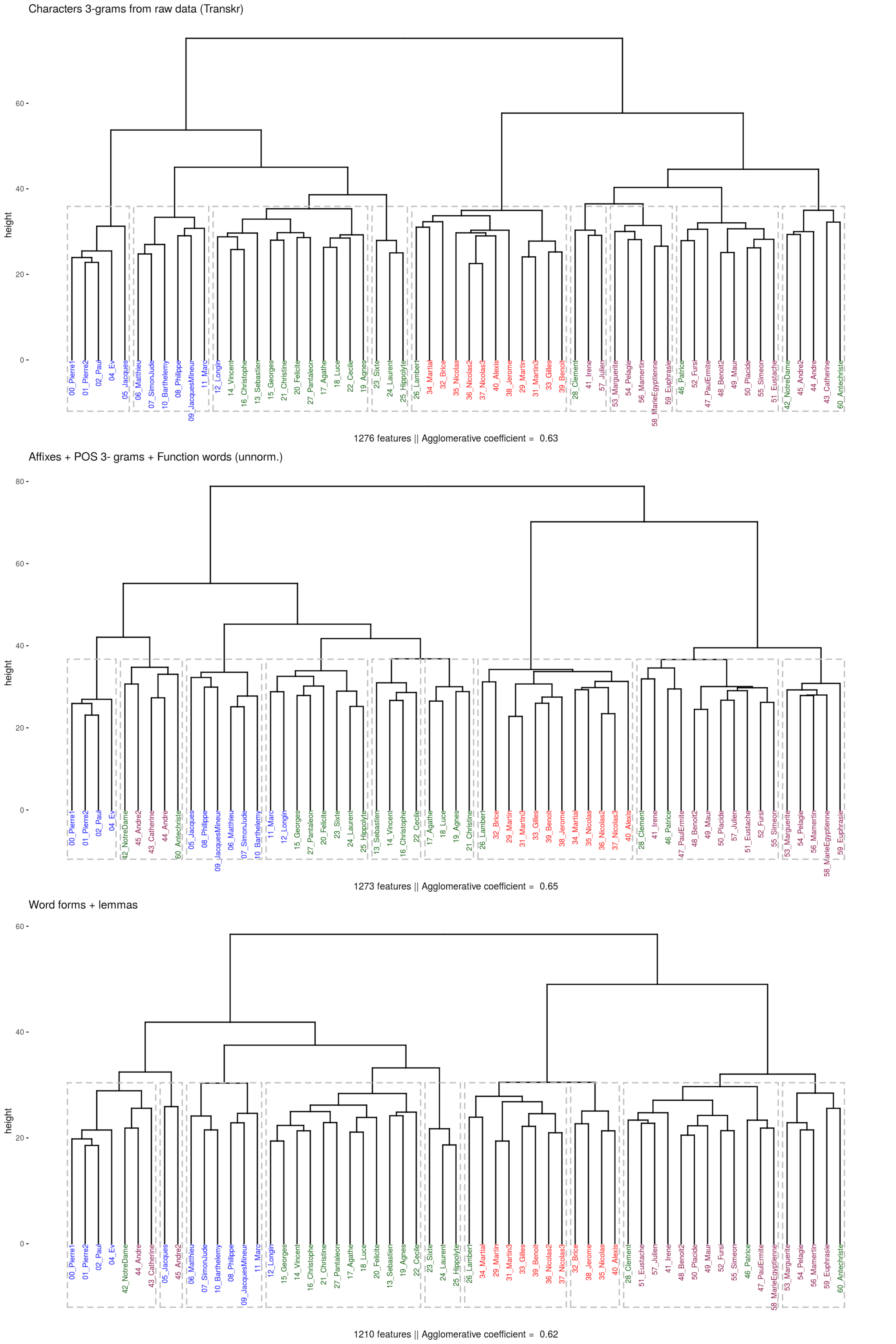}
    \caption{HC1: Main results from agglomerative hierarchical
clustering (Ward's method, Manhattan distance with z-scores and
vector-length euclidean normalisation) on the three feature sets (top,
character n-grams from HTR data; middle, functors; bottom, word forms
and lemmas).}
    \label{fig:7}
\end{figure}

\begin{figure}
    \centering
    \includegraphics[width=0.9\textwidth]{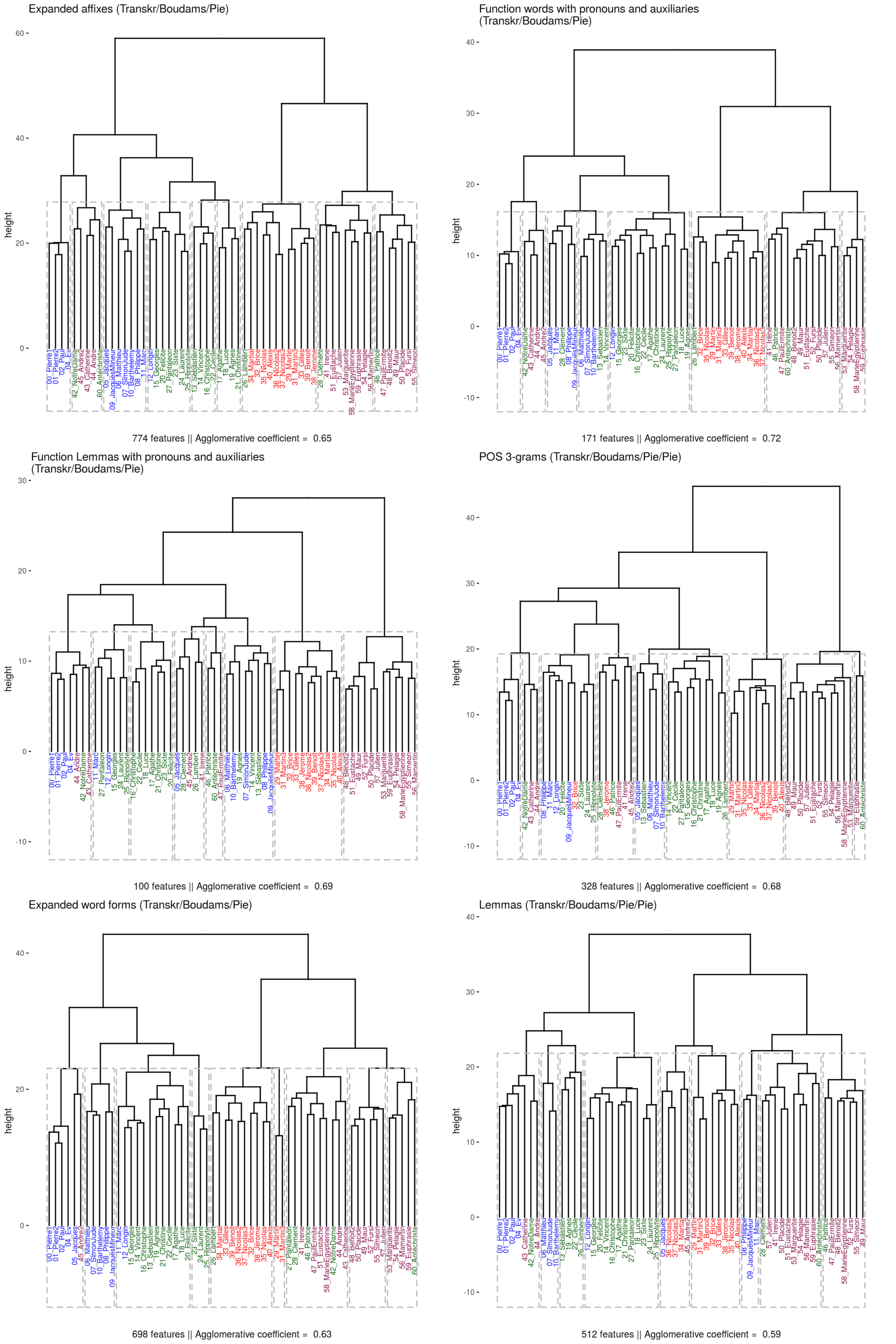}
    \caption{HC2: Supplementary results from agglomerative hierarchical
clustering (identical setup) on the components of feature sets 2 and 3}
    \label{fig:HC2}
\end{figure}

In order to check the robustness of our results, we give, for each
analysis shown in fig. 7 HC1 and fig. 8 HC2, four indicators:

\begin{itemize}
\item
  \emph{The number of analysed features and the agglomerative
  coefficient}, that, taken together, give an indication of the quality
  of the clustering;
\item
  The cluster purity of the groups (with \emph{k} = 5) as compared with
  Meyer's Hypothesis on A, B and C, and Wauchier's alleged texts;
\item
  The cluster purity of the groups when compared to our baseline
  (results on feature set 1).
\end{itemize}

These figures are given in table 7.

\begin{table}
\small
\centering
\begin{tabular}{l|cccc}
& N & AC & CP Meyer & CP Ref\tabularnewline \hline \hline
\textit{Feature Set 1} & & & &\tabularnewline
Raw 3-grams & 1276 & 0.63 & 0.90 & 1\tabularnewline \hline
\textit{Feature Set 2} & & & &\tabularnewline
Affixes & 774 & 0.65 & 0.85 & 0.92\tabularnewline
Function Words & 171 & 0.72 & 0.81 & 0.86\tabularnewline
Function Lemmas & 100 & 0.69 & 0.76 & 0.76\tabularnewline
POS3gr & 328 & 0.68 & 0.71 & 0.69\tabularnewline
Function words + POS 3-grams + Affixes & 1273 & 0.65 & 0.83 &
0.90\tabularnewline \hline
\textit{Feature set 3} & & & &\tabularnewline
Forms & 698 & 0.63 & 0.85 & 0.95\tabularnewline
Lemmas & 512 & 0.59 & 0.75 & 0.80\tabularnewline
Words + Lemmas & 1210 & 0.62 & 0.85 & 0.93\tabularnewline
\end{tabular}
\caption{Number of analysed features (N), agglomerative
coefficient (AC), cluster purity of the clusters (k = 5) when compared
to Meyer's hypothesis and Wauchier's known texts (CP Meyer), cluster
purity when compared to the baseline}
\label{table:7}
\end{table}

Eder  (2017) showed
that, whatever the variation of sample length, some texts were never
correctly attributed (at least when a given feature set is used) and
suggested to measure the diversity of attributions of individual texts -
what we call volatility - to help identify these cases when the authors
are not known. Following this, one could hypothesize that the presence
of classification resistant (or volatile) texts is to be expected in a
sufficiently large corpus.

To measure the volatility of any individual text, in the context of
unsupervised analysis, we wish to measure the stability or volatility of
its neighborhood. We devise a specific metric
$V_i$
 that aims to compute the volatility of
the neighborhood of a specific text $i$ in the groups of which he
is a member in all the clusterings performed. Let
$(G_j)_{j \in J, i \in G_j}$
be the
family of sets of which $i$ is a member, where $J$ is the
total number of clusterings performed. We can then construct a set
$X$, containing all unique texts
$\left\{x_a \dots x_n\right\} $ occurring in at least one
set of the family
$(G_j)$,
$X = \left\{x | \exists j \in J, x\in G_j \right\}$.
For each $x$, the family $(G_j)$ can be
split in two subfamilies, 
$(A_k)_{k \in J, x \in A_k}$
and $(B_l)_{l \in J, x \not\in B_k}$. We then
compute a global volatility index as follows:

\[
V_i = 
\frac{\sum_{a=1}^n
    \frac{%
    \mathbf{card}(A_k)_{k \in J, x_a \in A_k} - \mathbf{card}(B_l)_{l \in J, x_a \not\in B_l}
    }{%
    \mathbf{card}((G_j)_{j \in J})
    }}{%
\mathbf{card}(X)}
\]

Since we normalize it by the total number of elements in $X$, this
index is limited by $[-1;1]$, where -1 would indicate a perfect
volatility (all sets with no member in common) and 1 perfect stability
(all sets with the same members).

%\begin{table}
\begin{small} 
\centering
\begin{longtable}{l|ccc} \footnotesize %
Texts & N. words & V ref & V suppl\tabularnewline \hline \hline \endhead
11\_Ano\_Leg-A\_Ap\_NA\_Vie\_Marc & 1820 & -0.11 & -0.37\tabularnewline
05\_Ano\_Leg-A\_Ap\_NA\_Vie\_Jacques & 17920 & -0.05 &
-0.43\tabularnewline
42\_Ano\_Leg-B\_Vi\_NA\_Ass\_NotreDame & 3119 & 0 & -0.24\tabularnewline
43\_Ano\_Leg-C\_Vi\_NA\_Vie\_Catherine & 8877 & 0 & -0.24\tabularnewline
44\_Ano\_Leg-C\_Ap\_NA\_Vie\_Andre & 3118 & 0 & -0.24\tabularnewline
45\_Ano\_Leg-C\_Ap\_NA\_Pas\_Andre2 & 13315 & 0 & -0.44\tabularnewline
60\_Ano\_Leg-B\_NA\_NA\_NA\_Antechriste & 1485 & 0.25 &
-0.14\tabularnewline
00\_Ano\_Leg-A\_Ap\_Ev\_Dis\_Pierre1 & 6774 & 0.53 & 0.47\tabularnewline
01\_Ano\_Leg-A\_Ap\_NA\_Vie\_Pierre2 & 5527 & 0.53 & 0.47\tabularnewline
02\_Ano\_Leg-A\_Ap\_NA\_Pas\_Paul & 4798 & 0.53 & 0.47\tabularnewline
04\_Ano\_Leg-A\_Ap\_NA\_Vie\_Jean\_Ev & 4955 & 0.53 &
0.47\tabularnewline
10\_Ano\_Leg-A\_Ap\_NA\_Vie\_Barthelemy & 4360 & 0.71 &
-0.17\tabularnewline
28\_Ano\_Leg-B\_Ma\_Ho\_Vie\_Clement & 2544 & 0.71 &
-0.18\tabularnewline
41\_Ano\_Leg-C\_Vi\_NA\_Vie\_Irene & 3145 & 0.71 & -0.13\tabularnewline
46\_Ano\_Leg-B\_Co\_NA\_Pur\_Patrice & 7872 & 0.71 &
-0.13\tabularnewline
47\_Ano\_Leg-C\_Co\_er\_Vie\_PaulErmite & 3753 & 0.71 &
-0.13\tabularnewline
48\_Ano\_Leg-C\_Co\_ev\_Tra\_Benoit2 & 3234 & 0.71 & 0.36\tabularnewline
49\_Ano\_Leg-C\_NA\_NA\_Vie\_Maur & 6310 & 0.71 & 0.36\tabularnewline
50\_Ano\_Leg-C\_NA\_NA\_Vie\_Placide & 2783 & 0.71 & 0.36\tabularnewline
51\_Ano\_Leg-C\_Ma\_ho\_Vie\_Eustache & 3099 & 0.71 &
0.36\tabularnewline
52\_Ano\_Leg-C\_Co\_NA\_Vie\_Fursi & 2492 & 0.71 & 0.36\tabularnewline
53\_Ano\_Leg-C\_Vi\_NA\_Vie\_Marguerite & 1935 & 0.71 &
0.36\tabularnewline
54\_Ano\_Leg-C\_Vi\_NA\_Vie\_Pelagie & 1506 & 0.71 & 0.36\tabularnewline
55\_Ano\_Leg-C\_Co\_NA\_Vie\_Simeon & 2894 & 0.71 & 0.36\tabularnewline
56\_Ano\_Leg-C\_Co\_NA\_Vie\_Mamertin & 2202 & 0.71 &
0.36\tabularnewline
57\_Ano\_Leg-C\_Vi\_NA\_Vie\_Julien & 2766 & 0.71 & 0.36\tabularnewline
58\_Ano\_Leg-C\_Vi\_NA\_Vie\_MarieEgyptienne & 5529 & 0.71 &
0.36\tabularnewline
59\_Ano\_Leg-C\_Vi\_NA\_Vie\_Euphrasie & 1293 & 0.71 &
0.36\tabularnewline
06\_Ano\_Leg-A\_Ap\_NA\_Vie\_Matthieu & 6447 & 0.71 &
-0.17\tabularnewline
07\_Ano\_Leg-A\_Ap\_NA\_Vie\_SimonJude & 6784 & 0.71 &
-0.17\tabularnewline
08\_Ano\_Leg-A\_Ap\_NA\_Vie\_Philippe & 1014 & 0.71 &
-0.32\tabularnewline
09\_Ano\_Leg-A\_Ap\_NA\_Vie\_JacquesMineur & 1356 & 0.71 &
-0.32\tabularnewline
12\_Ano\_Leg-A\_Ma\_Ho\_Vie\_Longin & 2244 & 0.92 & 0.11\tabularnewline
13\_Ano\_Leg-B\_Ma\_Ho\_Vie\_Sebastien & 3539 & 0.92 &
-0.1\tabularnewline
14\_Ano\_Leg-B\_Ma\_Ho\_Vie\_Vincent & 4838 & 0.92 &
-0.05\tabularnewline
15\_Ano\_Leg-B\_Ma\_Ho\_Vie\_Georges & 4548 & 0.92 & 0.32\tabularnewline
16\_Ano\_Leg-B\_Ma\_Ho\_Vie\_Christophe & 9122 & 0.92 &
0.32\tabularnewline
17\_Ano\_Leg-B\_Ma\_Fe\_Vie\_Agathe & 3109 & 0.92 & 0.32\tabularnewline
18\_Ano\_Leg-B\_Ma\_Fe\_Vie\_Luce & 2366 & 0.92 & 0.32\tabularnewline
19\_Ano\_Leg-B\_Ma\_Fe\_Vie\_Agnes & 4177 & 0.92 & -0.07\tabularnewline
20\_Ano\_Leg-B\_Ma\_Fe\_Vie\_Felicite & 1676 & 0.92 &
0.11\tabularnewline
21\_Ano\_Leg-B\_Ma\_Fe\_Vie\_Christine & 7481 & 0.92 &
0.32\tabularnewline
22\_Ano\_Leg-B\_Ma\_Fe\_Vie\_Cecile & 6782 & 0.92 & 0.24\tabularnewline
23\_Ano\_Leg-B\_Ma\_Ho\_Vie\_Sixte & 1894 & 0.92 & 0.11\tabularnewline
24\_Ano\_Leg-B\_Ma\_Ho\_Vie\_Laurent & 3243 & 0.92 & 0.11\tabularnewline
25\_Ano\_Leg-B\_Ma\_Ho\_Vie\_Hippolyte & 2513 & 0.92 &
0.11\tabularnewline
27\_Ano\_Leg-B\_Ma\_Ho\_Vie\_Pantaleon & 6565 & 0.92 &
-0.28\tabularnewline
26\_Ano\_Leg-B\_Ma\_Ev\_Vie\_Lambert & 5247 & 1 & -0.27\tabularnewline
29\_Wau\_Leg-C\_Co\_Ev\_Vie\_Martin & 14432 & 1 & 0.64\tabularnewline
31\_Wau\_Leg-C\_Co\_Ev\_Dia\_Martin3 & 18971 & 1 & 0.64\tabularnewline
32\_Wau\_Leg-C\_Co\_Ev\_Vie\_Brice & 1385 & 1 & -0.04\tabularnewline
33\_Wau\_Leg-C\_Co\_Er\_Vie\_Gilles & 4415 & 1 & 0.64\tabularnewline
34\_Wau\_Leg-C\_Co\_Ev\_Vie\_Martial & 15255 & 1 & 0.64\tabularnewline
35\_Wau\_Leg-C\_Co\_Ev\_Vie\_Nicolas & 1960 & 1 & 0.64\tabularnewline
36\_Wau\_Leg-C\_Co\_Ev\_Mir\_Nicolas2 & 10473 & 1 & 0.64\tabularnewline
37\_Wau\_Leg-C\_Co\_Ev\_Tra\_Nicolas3 & 8379 & 1 & 0.64\tabularnewline
38\_Wau\_Leg-C\_Co\_Ev\_Vie\_Jerome & 2425 & 1 & -0.04\tabularnewline
39\_Wau\_Leg-C\_Co\_Ev\_Vie\_Benoit & 12792 & 1 & 0.64\tabularnewline
40\_Wau\_Leg-C\_Co\_Er\_Vie\_Alexis & 4103 & 1 & 0.64\tabularnewline
\caption{For each text, its number of words, and volatility index
(V) based on the three reference analyses (V Ref) or the supplementary
analyses (V Suppl). Texts are sorted on V Ref in increasing order}
\label{table:8}
\end{longtable}
%\end{table}
\end{small}

The results of this procedure are given in table 8. We notice that the
texts attributed to Wauchier are the least volatile, while there is a
small group of volatile texts achieving a score \textless{} 0.5.

Another indication yielded by this index is that volatility is not (or
almost not) due to variation in sample length (See Fig. 9). The small
relationship, on the edge of significance, between text length and
volatility that we observe when looking only at the supplementary
analyses disappears totally when we look at the reference analyses. This
could be an indication that the strategy we have adopted, of
concatenating several measurements to increase reliability on short
texts, is working.

\begin{figure}
    \centering
    \includegraphics[width=0.6\textwidth]{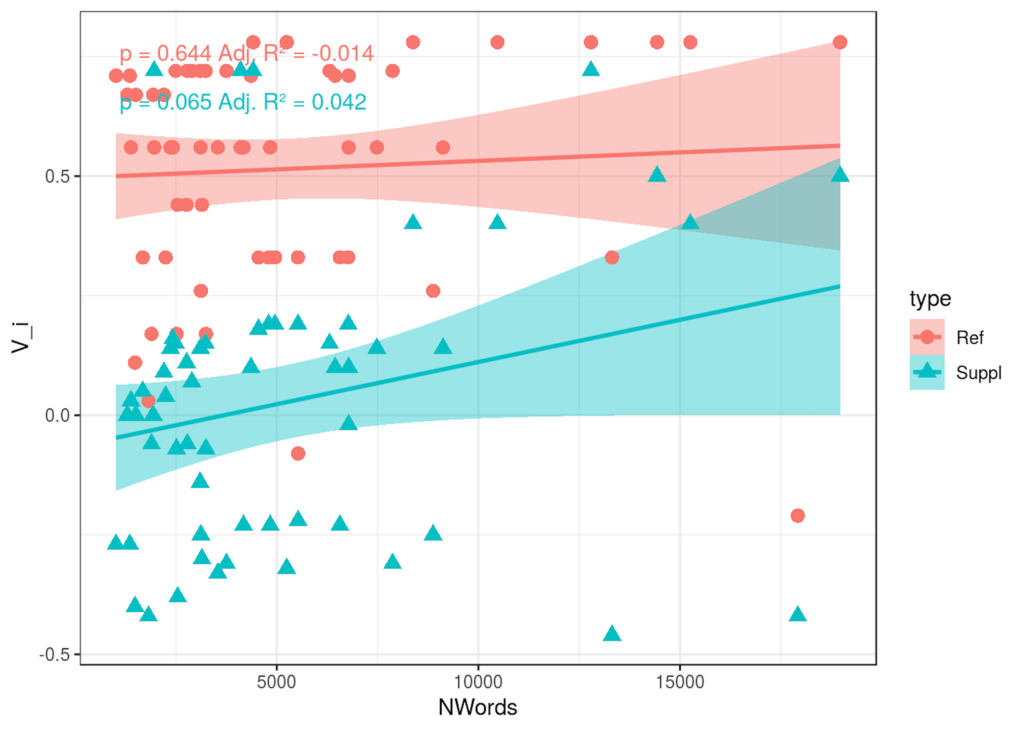}
    \caption{Scatterplot and regression lines for V \textasciitilde{}
NWords, for both reference analyses and supplementary analyses}
    \label{fig:9}
\end{figure}

\subsection{Controlling for pipeline
bias}\label{controlling-for-pipeline-bias}

To control for the presence of bias due to the training data used by the
pipeline, we perform the same set of analyses on the data obtained with
models trained on alternate data (TNAH corpus) or with different tools
(Kraken), and compare it with our analyses displayed above. The results
are displayed in table 9. Even though the models achieve quite different
accuracies, the results are not significantly different, and show a
particular stability on the three main analyses (mean CP of 0.95 and
0.92, table 9). The results based on a change of training corpus are
actually closer than the one obtained with the same corpus but a
different HTR software, though the difference remains small.

\begin{table}
\centering
\small
\begin{tabular}{l|cc}
& TNAH corpus & Kraken model\tabularnewline \hline \hline
Feature Set 1 & &\tabularnewline
Raw 3-grams & 0.93 & 0.90\tabularnewline
Feature Set 2 & &\tabularnewline
Affixes & 0.90 & 0.92\tabularnewline
Function Words & 0.83 & 0.73\tabularnewline
Function Lemmas & 0.78 & 0.71\tabularnewline
POS3gr & 0.81 & 0.88\tabularnewline
FW + POS 3-grams + Affixes & 0.98 & 0.92\tabularnewline
Feature set 3 & &\tabularnewline
Forms & 0.84 & 0.90\tabularnewline
Lemmas & 0.84 & 0.73\tabularnewline
Words + Lemmas & 0.95 & 0.95\tabularnewline
Geom. mean Main analyses & 0.95 & 0.92\tabularnewline
Geom. mean Suppl. analyses & 0.83 & 0.81\tabularnewline
Geom. mean all & 0.87 & 0.84\tabularnewline
\end{tabular}
\caption{Cluster purity of the analyses replicated using models
trained on the TNAH corpus or with Kraken, with regard with the analyses
presented in fig.7 HC1 and fig.8 HC2}
\label{tab9}
\end{table}

\section{Interpretation of the
results}\label{interpretation-of-the-results}

The manuscript fr. 412 allows us to control the results of our
approaches by checking the unity of the Wauchier de Denain
collection\footnote{The collection contains the nine
  following texts : \emph{Saint Martin Life, Sulpicius Severus Dialog,
  Saint Brice, Saint Gilles, Saint Martial, Saint Nicolas, Saint Jerome,
  Saint Benoit }and\emph{ Saint Alexis Lives.}} in the stylometric
trees. On the three reference analyses (see Fig.~7 HC1), the Wauchier
group, with the adjunction of the \emph{Life of saint Lambert}, is the
most clearly distinguished group. This same configuration, with or
without \emph{Lambert}, is also visible on all supplementary analyses,
except the ones based on POS 3‑grams and lemmas (see Fig.~8 HC2). These
two analyses also achieve low agglomerative coefficient, given their
number of features, and low cluster purity, both in comparison with
Meyer's classification and with our baseline; facts which advocates for
considering them as outliers, with low reliability.

Moreover, in their globality, the results seem to agree with Meyer's
hypothesis, with CP from 0.83 to 0.9 for the reference analysis (0.71 to
0.85 for the others, see Table 7). This is particularly obvious for our
baseline (see Fig.~7 HC1, top, reproduced here as Fig.10), that
represents the manuscript fr. 412 as a successive addition of
collections A, B and C, which appear in separated branches. This can
also be observed in the other trees, even if in a slightly noisier
fashion. For this reason, Paul Meyer's hypothesis seems to be confirmed
by our results. Nonetheless, they can be nuanced or made more accurate
in a few cases, as we will see.

\begin{figure}
    \centering
    \includegraphics[width=3.12639in,height=4.46389in]{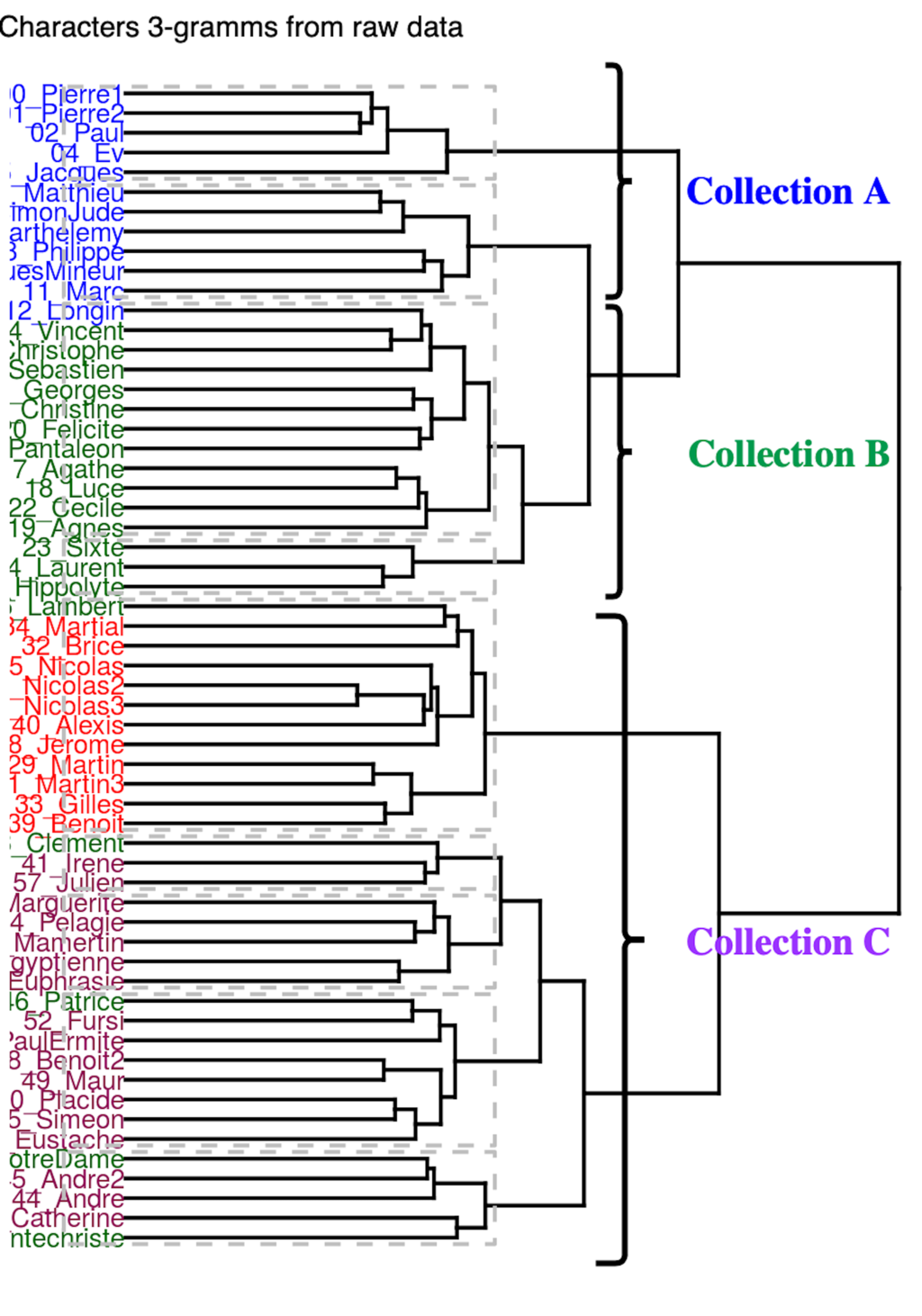}
    \caption{Results on char. 3-grams from raw HTR data, with
indication of the successive collections}
    \label{fig:my_label}
\end{figure}

\subsection{Volatile texts and exceptions to Meyer's
classification}\label{volatile-texts-and-exceptions-to-meyers-classification}

\subsubsection{Andrew-Catherine (43-45) and Assumption-Antichristi (42
and
60)}\label{andrew-catherine-43-45-and-assumption-antichristi-42-and-60}

A subgroup mixing texts from Meyer's B and C can be observed in our
results (see Fig. 7 HC1). It contains \emph{Saint Catherine Life},
\emph{Saint Andrew Life and Miracles}, \emph{Saint Andrew Passion }(n.
43-45\emph{)}, as well as the \emph{Assumption of Our Lady} and, but not
always, \emph{Antichristi }(n. 42 and
60)\footnote{Antechristi is only
  1346 words long, which is one of the smallest size of the corpora, and
  below the first quartile. One could hypothesize that the
   Jugement (61) could function with Antechristi, but this
  text was excluded because it is below the 1000 words limit.}.
In the trees, this subgroup is sometimes included in C, sometimes in
A.\emph{ }The A\emph{ssumption }and the \emph{Antichristi} were
identified by Paul Meyer as texts from collection B and the second one
was assumed to have certainly been published first in an autonomous way
with some others texts : \emph{Saint Patrice's Purgatory}, Julian and
Brendan Lives (Meyer, 1906:405).

\subsubsection{Clément and Patrice}\label{cluxe9ment-and-patrice}

Clément and Patrice are integrated in collection C (as opposed to
Meyer's B) in almost all analyses. A precise interpretation of this
remains to be found, but one can note that, Patrice has been supposed by
Paul Meyer as having been published first as part of a preexisting
\emph{libelli} and autonomously from collection B (or C) with
\emph{Antechristi}.

However, given the very small amount of texts erratically classified,
and given the pre-existing difficulties faced by Paul Meyer concerning
some of these, we do not expect this to contradict our former
conclusion. The apparent volatility of Marc and Jacques can partially be
disregarded, as they appear to only switch subgroups while remaining
inside A.

\subsection{Collection A}\label{collection-a}

Paul Meyer identified collection A as a group of twelve texts. It is
most apparent on the baseline analysis (see Fig. 7 HC1, top). From our
results, it can be refined in two subseries A1 and A2 (see Table 10) .
Thematically, this regroupment makes sense. On one side, we find major
apostles, creators of Christian Church, with two sequences of three and
two texts following each others. On the other side are minor apostles
which do not present sequences in theme or the likes.

\begin{table}
    \centering % 
    \small %
    \begin{tabular}{ll}
\textbf{series A1} & \textbf{series A2}\tabularnewline
\begin{minipage}[t]{0.47\columnwidth}\raggedright
\textbf{Sequence 1}

Dispute of Saint Peter and Saint Paul against Simon the magician (0)

Saint Peter Life and Passion (1)

Saint Paul Passion (2)

\textbf{Sequence 2}

{[}\emph{Martyrdom of Saint John in front of the Latin Door
(3)}{]}\textsuperscript{\footnotemark{}}

Saint John the Evangelist Life (4)\strut
\end{minipage}
\footnotetext{This text is amongst the one removed due to length below
  1000 words.} & \begin{minipage}[t]{0.47\columnwidth}\raggedright
Passion, translation and miracles of Saint James the Greater (5)

Saint Matthew Life (6)

Life of saint Simon and Jude (7)

Saint Philip Life (8)

Life of Saint James the Minor (9)

Saint Bartholomew Life (10)

Saint Marc Life (11)\strut
\end{minipage}\tabularnewline
\end{tabular}
\caption{Subseries of A}
\end{table}

To conclude about the collection A, in contrast to Meyer's hypothesis,
\emph{Saint Longin Life} is almost never grouped with any of the
aforementioned series, but is nearby, being clustered as the first
element of B.

\footnotetext{This text is amongst the one removed due to length below 1000 words.}

\subsection{Collection B}\label{collection-b}

Collection B contains nineteen texts, thematically centered around
martyrdoms. Of those, sixteen are sequentially found in MSS BnF fr. 412,
the three others being \emph{Our Lady, Antichristi, and Saint Patrice.
}The latest are nearly never grouped with the main body of B in the
trees\emph{. }

We can observe a strong group composed of twelve Lives: Christophe,
Agatha, Lucy, Agnes, Felicity, Christine, Cecile, Sixte, Laurent,
Hippolyte and Pantaleon Lives are clearly gathered (see Fig. 7 HC1). We
can add to them Georges, Vincent and Sebastien Lives, but saint Clement
Life is always missing. Thus, globally, the sequence of the manuscripts
is reflected in the classification.

Moreover, in all the selected trees, the \emph{Life of Saint Longin
}classified as A by Paul Meyer is gathered with texts from collection B.
Furthermore, thematically, \emph{Saint Longin's Life} isn't coherent
with a series of saint apostles, given the fact that he is a martyr. In
manuscript BnF fr. 412, given the order of the compilation, we can
considerate \emph{Saint Longin Life} as the last text of collection A or
as the first of collection B. Looking at the manuscripts tradition, in
fact this life is mixed with the apostle lives just once in the
manuscript BnF, nouv acq. fr. 23686, which happened to be the one Paul
Meyer used as his prime material for studying collection A. So regarding
our results and manuscript tradition, it seems more accurate to classify
this life to saint Martyrs within collection B.

In the light of this hypothetical classification, we can observe two
subcollections:

\begin{table}
    \centering 
    \small
\begin{tabular}{ll}
\textbf{Subseries B1} &\tabularnewline
\textbf{B1a - Men martyrs} & \textbf{B1b - Virgin martyrs}\tabularnewline
\begin{minipage}[t]{0.47\columnwidth}\raggedright
\textit{Saint Longin} (\textbf{Meyer's A -
12})

Saint Sebastien (13)

Saint Vincent (14)

Saint Christophe (16)\strut
\end{minipage} & \begin{minipage}[t]{0.47\columnwidth}\raggedright
Saint Agatha + Saint Luce (\emph{17-18})

Saint Agnes (19)

Saint Christine (21)

Saint Cecile (22)\strut
\end{minipage}\tabularnewline
\textbf{Subseries B2 - Roman martyrs?} &\tabularnewline
\begin{minipage}[t]{0.47\columnwidth}\raggedright
Saint Sixte (23)

Saint Laurent (24)

Saint Hyppolite (25)

Saint Pantaleon (27) ?\strut
\end{minipage} & \begin{minipage}[t]{0.47\columnwidth}\raggedright
\strut
\end{minipage}\tabularnewline
\end{tabular}
    \caption{Subcollection B1 of men and virgin martyrs and
subcollection B2}
\end{table}

A micro-series B1a is composed by \emph{Saint Longin, Saint Sebastien,
Saint Vincent, Saint Christopher Lives.}

Additionally, in B, a micro-series B1b of saint women Lives appears :
Saint Agatha, Saint Lucy, Saint Agnes, Saint Christine and saint Cecile.
Those Lives are about virgins and also close in the manuscript
tradition: the first three texts of the series are often gathered
together, as are the last three. There are also textual links between
them. One explanation for the proximity between Saint Agatha and Saint
Lucy can be the fact that the last seems to be in the continuation of
the first's story\footnote{In the manuscript tradition
  ( we check 18 manuscripts from \emph{Li Seint Confessor} tradition)
  there is a really strong link between Agatha and Lucy Lives which
  appear together in 11 manuscripts, against 7 witnesses where they
  appear separated in which only two manuscripts have both Lives.}.
Indeed, at the end of her story, Lucy defines herself as an heir of
Agatha:

\emph{Aussi com la cites de Cathenense est secorue et aidie par seinte
Agathe ma seror, aussi sera ceste citez aidie et socorue par moi, se uoz
auez foi et creance en nostre Signor}\footnote{Manuscript BnF fr. 412, fol. 72v.}.

Just as the city of Catania was rescued by the help of St. Agatha, my
sister, this city will be rescued by my help, if you have faith in our
Lord.

We can add that both of them come from Sicilia: Agatha from Catania and
Luce from Syracuse. There are also some links between the Lives of Saint
Lucy and Saint Agnes. The Life of Agnes starts at Rome where the Life of
Saint Lucy stopped and they both have to face the threat of a spurned
lover who wants to send them to a brothel. We can also note that saint
Christine, as saint Agatha, is one of the four patrons of Palermo in
Sicilia and that the thematic of the snatched breast, iconic for saint
Agatha, can also be found in saint Christine's Life. The reason behind
the adjonction of saint Cecile is more obscure: there might be a
redundant theme around family and conversion. Finally, we can add that
the Lives of this group are amongst the least volatile in our corpus
after the Wauchier de Denain collection (see Table 8).

However, both micro-series B1a and B1b can be grouped together as a
collection B1 in five of the selected trees. The rapprochement seems
logical from the point of view of literary construction because it
builds a collection with, on one side, five Lives of men martyrs, and on
the other side, six Lives of women martyrs. A stylometric study cannot
determine the order of apparition.

Finally, Paul Meyer, during his work about the different hagiographic
collections, has seen that, in the collection B, the series
Sixte-Laurent-Hippolyte was frequent in the manuscript tradition (Meyer,
1906:495). This reunion appears in our three analyses. Furthermore, the
dendrogram based on function words (see Fig. 8 HC2, top-right) links
Laurent and Hippolyte with Pantaleon. This addition isn't in
contradiction with the tradition: collection
G\footnote{Bruxelles, Bibliothèque royale, 9225, Paris,
  BnF, fr. 183, Paris, BnF, fr. 185.} contains them sequentially in
three of its four witnesses, and in collection C (three manuscripts)
saint Pantaleon's Life is only separated by Saint Lambert's Life from
the other ones. As such, it is possible that their gathering confirms a
predating series.

\subsection{Collection C}\label{collection-c}

Collection C contains twenty-two texts without any apparent major theme.
Collection C seems to have two major series, one constituted by Wauchier
de Denain's \emph{Seint Confessor}, and the other one containing all the
others texts.

\begin{table}
    \centering 
    \small
    \begin{tabular}{ll} 
\textbf{C1 -- Wauchier de Denain, }\emph{\textbf{Li Seint Confessor}}
&\tabularnewline
\begin{minipage}[t]{0.47\columnwidth}\raggedright
Saint Martin (29)

{[}Saint \emph{Martin 2 (30)}{]} \textsuperscript{\footnotemark{}}

Saint Martin 3 (31)

Saint Brice (32)

Saint Gilles (33)

Saint Martial (34)

Saint Nicolas (35)

Saint Nicolas 2 (36)

Saint Nicolas 3 (37)

Saint Jerome (38)

Saint Benoit (39)

Saint Alexis (40)\strut
\end{minipage}
\footnotetext{Wauchier's \emph{Martin 2} is amongst the texts removed
  due to their length being below 1000 words.} &
\begin{minipage}[t]{0.47\columnwidth}\raggedright
\strut
\end{minipage}\tabularnewline
Saint Lambert (Meyer's B) to attribute to Wauchier? &\tabularnewline
\textbf{C2 - Benedict and his disciples} &\tabularnewline
\begin{minipage}[t]{0.47\columnwidth}\raggedright
Translation of Saint Benoit (48)

Saint Maur (49)

Saint Placide (50)\strut
\end{minipage} & \begin{minipage}[t]{0.47\columnwidth}\raggedright
\strut
\end{minipage}\tabularnewline
\textbf{C3 - }&\tabularnewline
\textbf{C3a} - & \textbf{C3b} - \tabularnewline
\begin{minipage}[t]{0.47\columnwidth}\raggedright
Saint Marguerite (53)

Saint Pelagie (54)\\
Saint Euphrasie (59)\\
\emph{Saint Mary the egyptian (58)?}\strut
\end{minipage} & \begin{minipage}[t]{0.47\columnwidth}\raggedright
\emph{Saint Simeon (55) ?}

Saint Mamertin (56)\strut
\end{minipage}\tabularnewline
    \end{tabular}
    \caption{Subseries of C}
    \label{tab12}
\end{table}

First, we can see that the Lives of Saint Maur and Placide and
\emph{Saint Benoit's Translation}\footnote{
  The \emph{Translation of Saint Benoit} is not attributed to Wauchier
  neither by Paul Meyer nor by our stylometric analysis. Moreover, it is
  separated from the Life in MSS fr. 412 by 8 others texts.} form a
series C2. The \emph{translation} and the \emph{Life of saint Maur }are
grouped in our three analyses, and the \emph{Life of Saint Placide }is
also close to or part of the group. Those Lives have a thematic unity:
the translation of the body of Saint Benoit, followed by the Lives of
his disciples, Saint Maur and Saint Placide.

Another series C3 appears in all three analyses (see Fig. 7 HC1): Saint
Marguerite, Saint Pelagie, Saint Euphrasie and probably also \emph{The
Life of Saint Mary the Egyptian}. Surprisingly, we can extend this
series to a subseries C3b containing one, perhaps two, men's lives: the
\emph{Life of Saint Mamertin}, always present as well, while the
\emph{Life of Saint Simeon} is more punctually associated with this
group, when considering only function words or function lemmas.

Finally, this study has revealed an astonishing rapprochement between
\emph{Li seint Confessor} and the \emph{Life of saint Lambert}.
Normally, \emph{Saint Lambert's Life} is part of collection B, but
following our preceding analysis, Saint Lambert's doesn't fit in any
group of the collection. In fact, we have to look at some of the
supplementary analyses to find results where it is not associated with
Wauchier's works (POS 3-grams, function lemmas, and lemmas, Fig. 8 HC2),
all potentially influenced by the nature of the training corpus.
However, from a close reading perspective, it's difficult to affirm the
authorship given the fact that \emph{Saint Lambert Life} does not
possess any of the usual distinctive marks of Wauchier de Denain's style
such as verses in prose, signatures or vernacular translation of Latin
citations. Moreover, there is no references to Philippe of Namur, Jeanne
of Flanders or Roger squire of Lille, Wauchier de Denain's patrons.
There is also no evident Latin common substrate between the lives of
\emph{Li Seint Confessor }and \emph{Saint Lambert Life}. The only common
point which can be found is in the localization and the theme. Liege is
close to the Namur area\footnote{Indeed Wauchier de
  Denain comes form the North of France and worked under the patronage
  of the court of Flanders.}
 Saint Lambert is an
important saint, a bishop linked to power, having contact with Pepin,
king of the Franks. The chosen version is the one with a positive
representation of royal power. So, we are in the presence of a bishop,
ally of power, like Saint Martin or Saint Martial. On the other hand,
one possible hypothesis regarding a potential Wauchier's authorship is
that \emph{Saint Lambert Life} is an early text, where the author erases
himself and stays in the role of a simple translator. Consequently,
without any easy proof of classification, further study of \emph{Saint
Lambert Life}'s relationship with Wauchier's work should be done.

\section{Conclusion and further
research}\label{conclusion-and-further-research}

In this paper, we offer a complete pipeline to acquire and analyse
medieval data, using a hybrid approach of human and artificial
intelligence. Using machine learning, we are able to acquire and process
textual data from medieval manuscript images, and then submit it to a
stylometric analysis setup that seems able to deal with noisy data. The
application of this procedure offers perspectives for ancient or
medieval Cultural heritage data, and could be extended to other
material.

From a methodological and stylometric perspective, the challenge was, in
a context where supervised analysis was not an option, to deal with
short texts, whose data was noisy in two regards: first, because of the
noise generated during text recognition (HTR) and further processing
steps; second, because of the noise inherent to medieval data (spelling
variation, variants, etc.). From our observations, our baseline that
involved character n-grams with raw HTR data (already suggested as the
most adapted to noisy OCR or HTR data in previous studies) can still be
considered a very efficient procedure. Our attempts to suppress noise
due to spelling variation by using other types of features such as
lemmas or POS 3-grams, though offering alternative insights into the
data, do not seem yet able to surpass it significantly, perhaps because
of the cumulative error rates for each processing step. On the other
hand, concerning the shortness of the texts, our results seem to agree
with the notion that it is possible to analyse texts below 3000 words;
more specifically, by using less sparse features, such as characters
3-grams, or by concatenating different features, different views on the
same text, we seem to achieve stable results, independent of the
variation of sample length in our corpus.

From a thematic perspective, on the whole, our results confirm (or fail
to disprove) Meyer's hypothesis regarding the constitution of Old French
\emph{legendiers}. They also bring to light some new facts, such as
potential subseries that were not previously identified, and raise
questions about \emph{Saint Longin}'s life, that, we believe, can be
considered part of collection B instead of A, and the life of
\emph{Saint Lambert}, whose possible attribution to Wauchier de Denain
is deserving of further investigation.

Finally, we hope that our approach can motivate new investigations,
using computational humanities, on philological and historical holistic
hypotheses formulated in the 19th century, that still sometimes form the
basis of our understanding of the sources. By bringing together the work
of the founders of our fields, such as Paul Meyer, and novel
computational methods, we can hope to achieve progress in many areas,
and perhaps more specifically in those that are left out of the literary
canon envisioned by many close reading studies.

%\bibliographystyle{plain}
%\bibliography{references}

\section*{References}

  {\emph{\emph{\textbf{Argamon,
  S. and Levitan,
  S.}}}}{\emph{\emph{
  (2005). Measuring the Usefulness of Function Words for Authorship
  Attribution. In
  }}}{\emph{\emph{In
  Proceedings of the 2005 ACH/ALLC
  Conference}}}. \url{http://tomcat-stable.hcmc.uvic.ca:8080/ach/site/xhtml.xq?id=162}.

  \textbf{Cafiero, F. and Camps, J.-B.} (2019). Why Molière Most Likely
  Did Write His Plays. \emph{Science Advances}.

  {\emph{\emph{\textbf{Camps,
  J.-B.
  (ed.)}}}}{\emph{\emph{
  (2019).
  }}}{\emph{\emph{Geste:
  Un Corpus de Chansons de Geste, 2016-\ldots{}~
  }}}{\emph{\emph{(Version
  02)}}}{\emph{\emph{.
  Paris. \url{http://doi.org/10.5281/zenodo.2630574}.}}}

  {\emph{\emph{\textbf{Camps,
  J.-B. and Cafiero,
  F.}}}}{\emph{\emph{
  (2013). Setting Bounds in a Homogeneous Corpus: A Methodological Study
  Applied to Medieval Literature.
  }}}{\emph{\emph{Revue
  Des Nouvelles Technologies de l'information
  (RNTI)}}}{\emph{\emph{,
  }}}{\emph{\emph{\textbf{SHS-1}}}}{\emph{\emph{,
  pp. 55--84.}}}

  \textbf{Careri, M., et al.} (2001). \emph{Album de manuscrits français
  du XIIIe siècle}. Rome: Viella.

  {\textbf{Clérice, T.
  (2019)}}{ Evaluating
  Deep Learning Methods for Word Segmentation of Scripta Continua Texts
  in Old French and Latin.} \emph{Journal of Data Mining and Digital
  Humanities}, Episciences.org, 2020,
  \href{https://hal.archives-ouvertes.fr/hal-02154122v2}{\emph{https://hal.archives-ouvertes.fr/hal-02154122}}\href{https://hal.archives-ouvertes.fr/hal-02154122v2}{\emph{\emph{v2}}}

  {\textbf{Clérice, T.,
  Camps, J.-B., Pinche,
  A.}} (2019).
  Deucalion, Modèle Ancien Francais (0.2.0). Zenodo.
   \url{https://doi.org/10.5281/zenodo.3237455}

  \textbf{Douchet, S.} (Ed.) (2015). Wauchier de Denain, polygraphe du
  XIIIe siècle, Aix-en-Provence: Presses universitaires de Provence.

  {\emph{\emph{\textbf{Eder,
  M.}}}}{\emph{\emph{
  (2017). Short Samples in Authorship Attribution: A New Approach. In
  }}}{\emph{\emph{DH}}}. \url{https://dh2017.adho.org/abstracts/341/341.pdf.}
  (accessed 1 November 2019).

  {\emph{\emph{\textbf{Eder,
  M.}}}}{\emph{\emph{
  (2015). Does Size Matter? Authorship Attribution, Small Samples, Big
  Problem.
  }}}{\emph{\emph{Literary
  and Linguistic
  Computing}}}{\emph{\emph{,
  }}}{\emph{\emph{\textbf{30}}}}{\emph{\emph{(2),
  pp. 167--82. 10.1093/llc/fqt066.}}}

  \textbf{Eder, M. (}2013). Mind your corpus: systematic errors in
  authorship attribution. \emph{Literary and Linguistic Computing},
  28:603--614. \url{https://doi.org/10.1093/llc/fqt039}.

  \textbf{Evert, S., Proisl, T., Jannidis, F., Reger, I., Pielström, S.,
  Schöch, C., Vitt, T. (}2017). Understanding and explaining Delta
  measures for authorship attribution.
  {\emph{\emph{Digital
  Scholarship in the
  Humanities}}}{\emph{\emph{,
  }}}{\emph{\emph{\textbf{32}}}}{\emph{\emph{(suppl\_2),
  pp. ii4--ii16. 10.1093/llc/fqx023}}}.

  \textbf{Franzini, G., Kestemont, M., Rotari, G., Jander, M., Ochab,
  J.K., Franzini, E., Byszuk, J., Rybicki, J. (}2018). Attributing
  Authorship in the Noisy Digitized Correspondence of Jacob and Wilhelm
  Grimm.
  {\emph{\emph{Frontiers
  in Digital
  Humanities}}}{\emph{\emph{,
  }}}{\emph{\emph{\textbf{5}}}}{\emph{\emph{.
  10.3389/fdigh.2018.00}}}.

  \textbf{Garz, A. }\emph{\textbf{et al.}}\textbf{ (}2012)
  `Binarization-Free Text Line Segmentation for Historical Documents
  Based on Interest Point Clustering', in \emph{2012 10th IAPR
  International Workshop on Document Analysis Systems}. \emph{2012 10th
  IAPR International Workshop on Document Analysis Systems}, pp. 95--99.
  doi:
  \href{https://doi.org/10.1109/DAS.2012.23}{\emph{\emph{10.1109/DAS.2012.23}}}.

  {\emph{\emph{\textbf{Gómez-Adorno,
  H. et
  al.}}}}{\emph{\emph{
  }}}{\emph{\emph{(2018).
  Document Embeddings Learned on Various Types of N-Grams for
  Cross-Topic Authorship Attribution.
  }}}{\emph{\emph{Computing}}}{\emph{\emph{,
  }}}{\emph{\emph{\textbf{100}}}}{\emph{\emph{(7),
  pp. 741--756.}}}

  \textbf{Ing, L.} (in progress). \emph{Disparitions Lexicales En
  Diachronie: Traitements Automatiques Sur Le Lancelot En Prose}. {[}PhD
  Thesis Thesis{]}. Paris: École Nationale Des Chartes.
  \url{http://www.theses.fr/s221114} (accessed 1 November 2019).

  \textbf{Jannidis, F. et al.} (2015). Improving Burrows' Delta-An
  Empirical Evaluation of Text Distance Measures. In \emph{Digital
  Humanities Conference}.
  \url{https://www.researchgate.net/profile/Steffen_Pielstroem/publication/280086768_Improving_Burrows'_Delta_-_An_empirical_evaluation_of_text_distance_measures/links/573ad8ae08ae9f741b2d3d40.pdf} (accessed 23 May 2017).

  {\emph{\emph{\textbf{Kestemont,
  M.}}}}{\emph{\emph{
  (2014). Function Words in Authorship Attribution. From Black Magic to
  Theory? In
  }}}{\emph{\emph{Proceedings
  of the 3rd Workshop on Computational Linguistics for Literature
  (CLFL)}}}{\emph{\emph{.
  pp. 59--66.}}}

  {\emph{\emph{\textbf{Koppel,
  M., Schler, J. and Argamon,
  S.}}}}{\emph{\emph{
  (2009). Computational Methods in Authorship Attribution.
  }}}{\emph{\emph{Journal
  of the American Society for Information Science \&
  Technology}}}{\emph{\emph{,
  }}}{\emph{\emph{60}}}{\emph{\emph{(1),
  pp. 9--26.
  }}}{\emph{\emph{10.1002/asi.20961.}}}

  \textbf{Kunstmann, P. (ed.)} (2009). \emph{Chrétien de Troyes: Cligès,
  Erec, Lancelot, Perceval, Yvain -\/- Manuscrit P (BnF Fr. 794)}.
  \url{http://www.atilf.fr/dect}.

  \textbf{Manjavacas,
  E., Kádár, Á. and Kestemont,
  M.}
  (2019). Improving Lemmatization of Non-Standard Languages with Joint
  Learning.
  {\emph{\emph{ArXiv:1903.06939
  {[}Cs{]}}}}. \url{http://arxiv.org/abs/1903.06939}
  (accessed 23 October 2019).

  \textbf{Manjavacas, E., Kestemont, M., Clérice, T.} (2019).
  emanjavacas/pie v0.2.3. Zenodo.
  \href{https://doi.org/10.5281/zenodo.2654987}{\emph{\emph{https://doi.org/10.5281/zenodo.2654987}}}

  \textbf{Meyer, P.} (1906) `Légendes hagiographiques en français', in
  \emph{Histoire littéraire de la France}. Imprimerie nationale. Paris,
  pp. 328--458.

  {\emph{\emph{\textbf{Moisl,
  H.}}}}{\emph{\emph{
  (2011). Finding the Minimum Document Length for Reliable Clustering of
  Multi-Document Natural Language Corpora.
  }}}{\emph{\emph{Journal
  of Quantitative
  Linguistics}}}{\emph{\emph{,
  }}}{\emph{\emph{\textbf{18}}}}{\emph{\emph{(1),
  pp. 23--52. 10.1080/09296174.2011.533588.}}}

  \textbf{Olivier-Martin (F.), Duval, F., Ing, L.} (2018). \emph{Les
  Institutes de Justinien en français, Paris, 1935}, éd. revue par F.
  Duval, lemmatisée par F. Duval et L. Ing.

  \textbf{Perrot, J.-P.} (1992) \emph{Le passionnaire français au Moyen
  Âge}. Genève, Suisse: Droz.

  \textbf{Philippart, G.} (1977) \emph{Les Légendiers latins et autres
  manuscrits hagiographiques}. Turnhout: Brépols.

  {\emph{\emph{\textbf{Pinche,
  A. et
  al.}}}}{\emph{\emph{
  (2019). Chartes-TNAH/digital-edition: 2019. 
  \href{http://dx.doi.org/10.5281/zenodo.3522144}{10.5281/zenodo.3522144}.}}}

  \textbf{Pinche, A.} (En cours). \emph{Edition Nativement Numérique Des
  Oeuvres Hagiographiques `Li Seint Confessor' de Wauchier de Denain
  d'après Le Manuscrit 412 de La Bibliothèque Nationale de France.}
  \emph{{[}PhD Thesis Thesis{]} Lyon:Université} Lyon
  3. \url{http://www.theses.fr/s150996} (accessed 9 May 2017).

  {\emph{\emph{\textbf{Sapkota,
  U. et
  al.}}}}{\emph{\emph{
  (2015). Not All Character N-Grams Are Created Equal: A Study in
  Authorship Attribution. In
  }}}{\emph{\emph{Proceedings
  of the 2015 Conference of the North American Chapter of the
  Association for Computational Linguistics: Human Language
  Technologies}}}{\emph{\emph{.
  pp. 93--102.}}}

  {\emph{\emph{\textbf{Stamatatos,
  E.}}}}{\emph{\emph{
  (2013). On the Robustness of Authorship Attribution Based on Character
  N-Gram Features.
  }}}{\emph{\emph{Journal
  of Law and
  Policy}}}{\emph{\emph{,
  }}}{\emph{\emph{\textbf{21}}}}{\emph{\emph{(2),
  pp. 421--439.}}}

  \textbf{Stutzmann, D. et al.} (2013\emph{). Ontology Research, Image
  Features, Letterform Analysis on Multilingual Medieval Scripts --
  ORIFLAMMS}

  {\emph{\emph{\textbf{Stutzmann,
  D.}}}}{\emph{\emph{
  (2019). Words as Graphic and Linguistic Structures: Word Spacing in
  Psalm 101 Domine Exaudi Orationem Meam (11th-15th c.).}}}

  {\emph{\emph{\textbf{Ward
  Jr, J.
  H.}}}}{\emph{\emph{
  (1963). Hierarchical Grouping to Optimize an Objective Function.
  }}}{\emph{\emph{Journal
  of the American Statistical
  Association}}}{\emph{\emph{,
  }}}{\emph{\emph{\textbf{58}}}}{\emph{\emph{(301),
  pp. 236--244. 10.2307/2282967.}}}

  \textbf{Wauchier de Denain} (1999). \emph{La vie de Mon Signeur Seint
  Nicholas le Beneoit confessor}. J. J. Thompson (ed.), Genève~:Droz.

\end{document}